\documentclass[10pt,twocolumn,letterpaper]{article}

\usepackage{cvpr}              

\usepackage{graphicx}
\usepackage{amsmath}
\usepackage{amssymb}
\usepackage{booktabs}
\usepackage{multirow}
\usepackage[ruled,vlined]{algorithm2e}
\usepackage{algorithmic}
\usepackage[accsupp]{axessibility} 

\usepackage{caption}
\usepackage{xcolor}
\definecolor{commentcolor}{RGB}{110,154,155}   
\newcommand{\PyComment}[1]{\ttfamily\textcolor{commentcolor}{\# #1}}  
\newcommand{\PyCode}[1]{\ttfamily\textcolor{black}{#1}} 

\usepackage[pagebackref,breaklinks,colorlinks]{hyperref}

\usepackage[capitalize]{cleveref}
\crefname{section}{Sec.}{Secs.}
\Crefname{section}{Section}{Sections}
\Crefname{table}{Table}{Tables}
\crefname{table}{Tab.}{Tabs.}

\def\cvprPaperID{6340} 
\def\confName{CVPR}
\def\confYear{2022}

\begin{document}

\title{BatchFormer: Learning to Explore Sample Relationships for Robust Representation Learning}
\author{Zhi Hou$^1$, Baosheng Yu$^1$, Dacheng Tao$^{2,1}$ \\
$^1$ School of Computer Science, Faculty of Engineering, The University of Sydney, Australia \\
$^2$ JD Explore Academy, China \\
{\tt\small zhou9878@uni.sydney.edu.au, baosheng.yu@sydney.edu.au, dacheng.tao@gmail.com}
}
\maketitle

\begin{abstract}
Despite the success of deep neural networks, there are still many challenges in deep representation learning due to the data scarcity issues such as data imbalance, unseen distribution, and domain shift. To address the above-mentioned issues, a variety of methods have been devised to explore the sample relationships in a vanilla way (i.e., from the perspectives of either the input or the loss function), failing to explore the internal structure of deep neural networks for learning with sample relationships. Inspired by this, we propose to enable deep neural networks themselves with the ability to learn the sample relationships from each mini-batch. Specifically, we introduce a batch transformer module or BatchFormer, which is then applied into the batch dimension of each mini-batch to implicitly explore sample relationships during training. By doing this, the proposed method enables the collaboration of different samples, e.g., the head-class samples can also contribute to the learning of the tail classes for long-tailed recognition. Furthermore, to mitigate the gap between training and testing, we
share the classifier between with or without the BatchFormer during training, which can thus be removed during testing. We perform extensive experiments on over ten datasets and the proposed method achieves significant improvements on different data scarcity applications without any bells and whistles, including the tasks of long-tailed recognition, compositional zero-shot learning, domain generalization, and contrastive learning. Code is made publicly available at \url{https://github.com/zhihou7/BatchFormer}.
\end{abstract}

\begin{figure}[!ht]
    \centering
    \includegraphics[width=.9\linewidth]{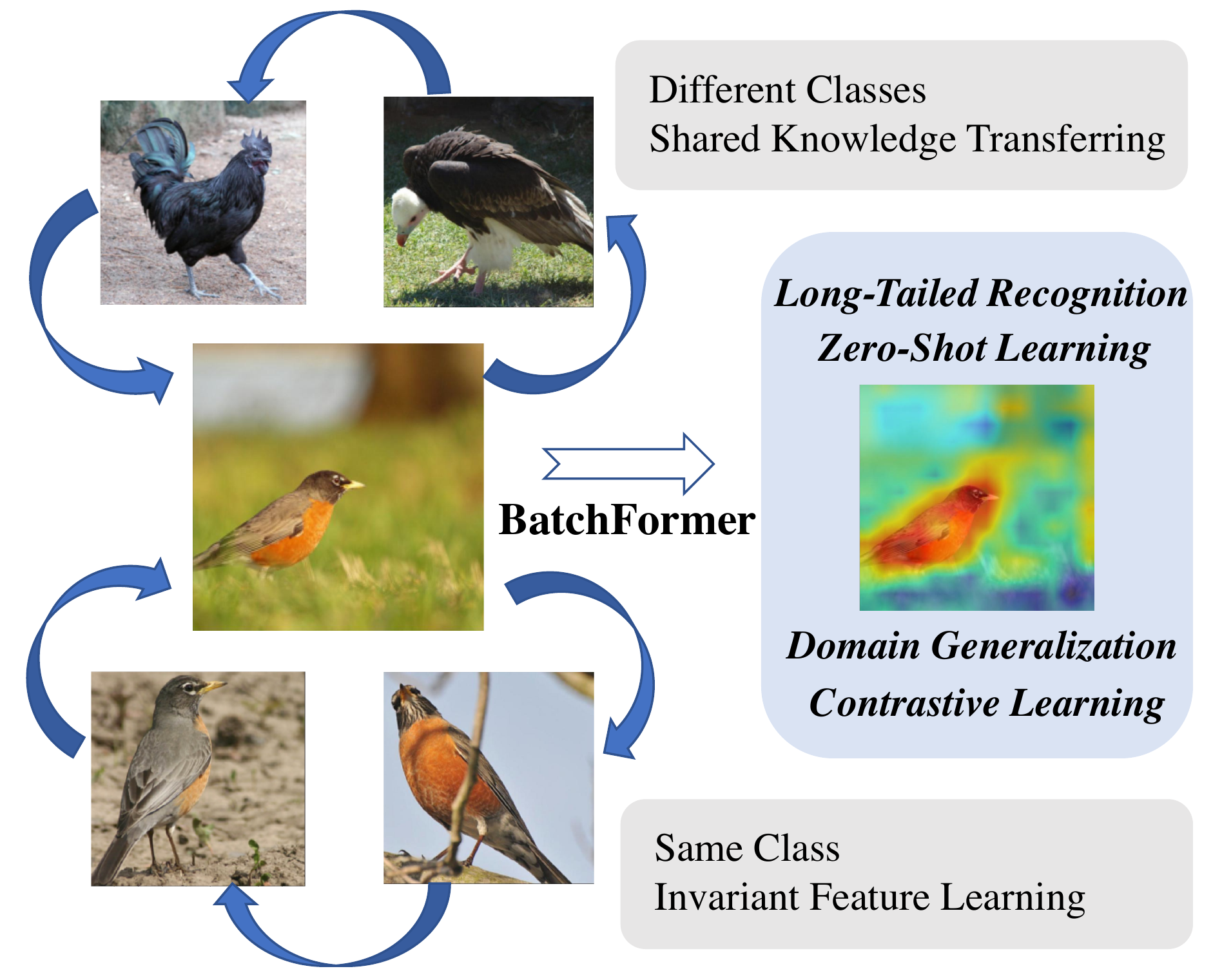}
    \caption{An illustration of the relationships between different images. Specifically, similar classes tend to share similar parts (\eg, cock, robin, vulture share body shape and claw shape) and transferable augmentation (\eg, angles). Therefore, transferring shared knowledge from head/seen classes to tail/unseen classes can facilitate long-tailed/zero-shot learning. In addition, exploring the invariant features between images belonging to the same class is also helpful for robust representation learning with a few samples.}
    \label{fig:illu}
\end{figure}

\section{Introduction}

Despite the great success of deep neural networks for representation learning~\cite{He2015, he2019moco}, it heavily relies on collecting large-scale training data samples, which turns out to be non-trivial in real-world applications. Therefore, how to form robust deep representation learning under the data scarcity by exploring the sample relationships has received a lot of attention from the community, especially for the tasks without good training data to guarantee the generalization, such as long-tailed recognition~\cite{wang2020long}, zero-shot learning~\cite{naeem2021learning}, and domain generalization~\cite{cha2021swad}. However, it still remains a great challenge to find a unified, flexible, and powerful way to explore the sample relationships for robust representation learning. An intuitive example revealing the effectiveness of sample relationships is shown in Figure~\ref{fig:illu}.




Among recent data scarcity learning methods, sample relationships have been intensively explored using an explicit scheme from either regularization~\cite{shankar2018generalizing, zhang2018mixup, li2021metasaug,ghiasi2021simple} or knowledge transfer~\cite{wang2017learning, wang2021rsg,zhu2020inflated}. Specifically, a simple yet very effective way is to directly generate new data samples/domains from existing training data~\cite{li2021metasaug}, such as mixup~\cite{zhang2018mixup},  copy-paste~\cite{ghiasi2021simple}, crossgrad~\cite{shankar2018generalizing}, and compositional learning~\cite{kato2018compositional, hou2020visual, atzmon2020causal, naeem2021learning}. Another way is to transfer knowledge between data samples, e.g., 1) transferring meta knowledge between head and tail classes for long-tailed recognition~\cite{wang2017learning, liu2019openlongtailrecognition}; 2) transferring the knowledge from seen classes to unseen classes for zero-shot learning~\cite{xian2018zero,naeem2021learning}; and 3)
transferring invariant knowledge for domain generalization~\cite{peng2019domain, arjovsky2019invariant, huang2020self}. However, the above-mentioned methods explore sample relationships from either the input or output of deep neural networks, failing to enable deep neural networks themselves with the ability to explore sample relationships, i.e., there is no interaction from the view of batch dimension.

Enabling the learning on the batch dimension is not easy for deep neural networks due to the training and inference gap, i.e., we do not always have a mini-batch of data samples during testing. For example, batch normalization requires to always keep mini-batch training statistics, where the running mean and variance are then used to normalize testing samples~\cite{ioffe2015batch}. Another example uses the feature memory to keep category centers during training, which is then used to enhance the tail/unseen categories during testing~\cite{liu2019openlongtailrecognition, zhu2020inflated}. Therefore, to explore sample relationships for robust representation learning, we propose to empower the deep neural networks with structural advances for sample relationship learning. Specifically, we try to capture and model the sample relationships in each mini-batch of training data samples by introducing a transformer into the batch dimension, and we refer to it as the Batch Transformer or BatchFormer. Furthermore, to mitigate the gap between training and testing, we utilize a shared classifier before and after the BatchFormer module to enforce the batch-invariant learning. By doing this, the BatchFormer module is only required during training, i.e., we do not need to change the inference structures of deep neural networks.

From the perspective of optimization, BatchFormer enables the information propagation of all features of the mini-batch samples. Therefore, all samples can contribute to the learning on any object categories, and the insight might be that this implicitly enriches current training samples with hallucinated features from the whole mini-batch (\eg, the shared parts between two categories). For example, in long-tailed recognition, the hallucinated features may improve the feature space of the tail classes. Meanwhile, the loss function also emphasizes on the rare classes via propagating larger gradients of rare classes on other features in the mini-batch. In particular, we also find that BatchFormer brings two obvious changes to the learned representation, \ie, it effectively facilitates the deep model to learn 1) comprehensive representations by focusing on almost all different parts of the object; and 2) invariant representations by focusing on the object itself rather than the complex background cues (See more empirical evaluations in appendix).

In this paper, our main contributions can be summarized as follow: 1) we propose to explore sample relationships from the perspective of the internal structure of deep neural networks; 2) we devise a simple yet effective module termed as BatchFormer, which is a plug-and-play module to explore sample relationships in each mini-batch; and 3) without any bells and whistles, we perform extensive experiments to demonstrate the effectiveness of BatchFormer in a variety of visual recognition tasks, including long-tailed recognition, zero-shot learning, domain generalization, and self-supervised representation learning.

\begin{figure*}[!ht]
    \centering
    \includegraphics[width=0.9\linewidth]{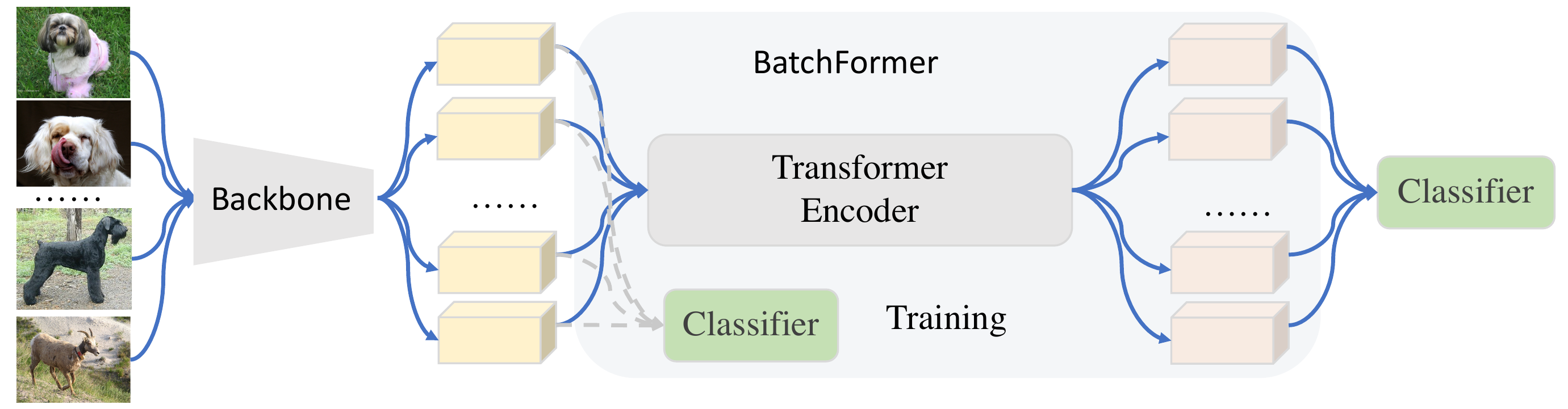}
    \caption{The main framework of representation learning with the proposed BatchFormer. Specifically, we apply a BatchFormer module between the feature extractor (\eg, ResNet) and the classifier layer to explore the samples relationships. Furthermore, with a shared classifier before and after the BatchFormer during training for batch-invariant learning, we can then remove BatchFormer during testing.}
    \label{fig:struct}
\end{figure*}

\section{Related Work}

{\bf Sample Relationship.}
There are diverse and firm relationships among different samples, which have been widely used via various types of strategies~\cite{zhang2018mixup, liu2018learning, hou2019cross, mondal2021mini}. Zhang \etal~\cite{zhang2018mixup} propose to regularize the network to favor simple linear behavior in-between training examples with mixup. Mixup~\cite{zhang2018mixup} merely considers the linear transformation between examples, while we investigate the relationship among examples in a non-linear way. The compositionality of samples has inspired massive approaches to improve the generalization of few-shot and zero-shot learning~\cite{tokmakov2019learning, he2021compas, hou2020visual, naeem2021learning}, where the shared parts/attributes among different samples have been explored via prior label relationship knowledge. There are also some approaches via investigating sample/class relationships to conduct transductive inference~\cite{liu2018learning, hou2019cross, liu2019openlongtailrecognition, mondal2021mini}, \eg, transductive few-shot classification~\cite{liu2018learning}, meta embedding~\cite{liu2019openlongtailrecognition, zhu2020inflated}. However, those approaches require to conduct inference with multiple samples (\eg, query set, or bank features). Meanwhile, massive domain generalization methods~\cite{peng2019domain, arjovsky2019invariant, huang2020self} aims to find casual/invariant representations across domains, which we think internally utilizes the relationship among samples of the same class but different domains. However, those techniques are diverse and complex. By contrast, we propose to investigate relationships for robust representation learning in a simple yet effective way, and thus benefit those challenging tasks simultaneously.




\textbf{Data Scarcity Learning}.
Learning with imperfect training data has turned out to be very challenging, which has been explored in a variety of data scarcity tasks. A very important problem is long-tailed recognition, where
the data from different classes usually exhibit a long-tailed distribution: a large portion of classes have very few instances. Current long-tailed approaches can be roughly categorized into distribution re-balancing methods (\eg, re-sampling~\cite{chawla2002smote, han2005borderline, he2009learning}, re-weighting~\cite{byrd2019effect, cui2019class, tan2020equalization,zhong2021improving, jamal2020rethinking}), ensemble of diverse branches/experts~\cite{zhou2020bbn, wang2020long, cai2021ace}, and knowledge transfer~\cite{wang2017learning, liu2019openlongtailrecognition, zhu2020inflated, he2021distilling, wang2021rsg}.
Many knowledge transfer approaches have been developed for long-tailed recognition via meta learning~\cite{wang2017learning}, memory features~\cite{liu2019openlongtailrecognition, zhu2020inflated}, and virtual data generation~\cite{he2021distilling, wang2021rsg, hou2021fcl}. Nevertheless, those works usually depend on complex memory bank~\cite{liu2019openlongtailrecognition, zhu2020inflated}, or complex learning strategies~\cite{wang2017learning} or additional distillation step~\cite{he2021distilling}.
Another data scarcity tasks include zero-shot learning and domain generalization~\cite{blanchard2011generalizing, zhou2021domain, wang2021generalizing}, which require to recognize new categories with unseen training data. Specifically, zero-shot learning aims to recognize unseen classes that do not training samples, while domain generalization targets at generalizing classes of seen domains to unseen domains. Current zero-shot learning approaches ~\cite{lampert2009learning, palatucci2009zero, xian2018zero, wang2019survey} usually transfer knowledge of seen classes to unseen classes via modern techniques (\eg, graph network~\cite{naeem2021learning}, data generalization~\cite{zhu2018generative}, and compositional learning~\cite{naeem2021learning, hou2020visual}) for unseen classes recognition. Recently, compositional zero-shot learning~\cite{kato2018compositional, hou2020visual, atzmon2020causal, naeem2021learning} have been widely explored in different tasks, we thus mainly evaluate the proposed BatchFormer on compositional zero-shot learning~\cite{naeem2021learning}. Domain generalization techniques usually include data augmentation~\cite{zhou2020learning, zhou2021domain}, meta learning~\cite{balaji2018metareg, zhao2021learning}, and disentangled/invariant representation learning~\cite{peng2019domain, arjovsky2019invariant, chattopadhyay2020learning, piratla2020efficient}.  BatchFormer facilitates invariant representation learning, and thus illustrates effectiveness on multiple datasets.



{\bf Contrastive Self-Supervised Learning.}
Contrastive Learning~\cite{hadsell2006dimensionality} has increasingly achieved great success in computer vision for self-supervised learning, \eg, \cite{oord2018representation}, \cite{hjelm2018learning}, \cite{he2019moco, chen2020mocov2, chen2021mocov3} and \cite{chen2020simple}. Contrastive Learning aims to learn representations that attract similar samples and dispel different samples, where the similar samples and different samples are known according to some priors. However, we implicitly mine the relationships with Transformer Encoder network. Specifically, we also demonstrates the effectiveness of BatchFormer in contrastive learning, \eg, MoCo-v2~\cite{chen2020mocov2} and MoCo-v3~\cite{chen2021mocov3}.

\section{Method}

In this section, we first provide an overview of the main deep representation learning framework with the proposed BatchFormer module. We then introduce the proposed BatchFormer module in detail, including the encoder and the shared classifier. Lastly, we discuss the insights behind the proposed BatchFormer from the view of gradient flow.

\subsection{Overview}

The relationships between different samples are various and complex, while previous approaches have explored sample relationships using a straightforward way, such as the joint manipulation of different input images and the knowledge transfer using meta embedding or loss functions. Among existing methods, sample relationships are usually required to be defined in an explicit scheme before they can be used for learning, thus failing to automatically learn sample relationships for representation learning.

We consider sample relationships from a learning perspective, i.e., we aim to enable deep neural networks themselves with the ability to learn sample relationships from each mini-batch sample during the end-to-end deep representation learning. The main deep representation learning framework with the proposed new module is shown in Figure~\ref{fig:struct}. Specifically, a backbone network is first used to learn representations for individual data samples, i.e., there is no interaction between different samples in each mini-batch. After this, a new module is introduced to model the relationships between different samples by using the cross-attention mechanism in transformer, and we thus referred to it as Batch Transformer or BatchFormer module. The output of BatchFormer is then used as the input of the final classifier. To fulfill the gap between training and testing, we also utilize an auxiliary classifier before the BatchFormer module, i.e., by sharing the weights between the final classifier and the auxiliary classifier, we are able to transfer the knowledge learned from sample relationships to the backbone and auxiliary classifier. Therefore, during testing we can remove the BatchFormer and directly use the auxiliary classifier for classification.


\begin{algorithm}[!ht]
  \algsetup{linenosize=\tiny}
  \scriptsize
\SetAlFnt{\tiny}
\SetAlCapFnt{\small}
\SetAlCapNameFnt{\small}
\SetAlgoLined
\PyCode{def BatchFormer(x, y, encoder, is\_training):} \\
\Indp
\PyComment{x: input features with the shape [N, C]} \\
\PyComment{encoder: TransformerEncoderLayer(C,4,C,0.5)} \\
\PyCode{if not is\_training:}\\
    \Indp
    return x, y \\
    \Indm
\PyCode{pre\_x = x} \\
\PyCode{x = encoder(x.unsqueeze(1)).squeeze(1)} \\
\PyCode{x = torch.cat([pre\_x, x], dim=0)} \\
\PyCode{y = torch.cat([y, y], dim=0)} \\
\PyCode{return x, y} \\
\Indm 
\caption{Pytorch Code of BatchFormer.}
\label{alg}
\end{algorithm}

\subsection{BatchFormer}

In this subsection, we introduce the detailed structures of the proposed BatchFormer. Specifically, the proposed BatchFormer module utilizes a stack of multiple transformer encoder layers to model the relationships between different samples.

\textbf{Transformer Encoder}. The transformer encoder includes multihead self-attention (MSA) and MLP blocks. A Layernorm (LN) is used after each block. Let $X \in R^{N\times C}$ denote a sequence of input features, where $N$ is the length of the sequence and $C$ indicates the dimension of input features. We then have the output of the transformer encoder as follows,
\begin{align}
\label{eq:msa}
\hat{X}_{l} &= LN(MSA(X_{l-1}) + X_{l-1}),\\
 X_l &= LN(MLP(\hat{X}_l) + \hat{X}_{l}),
\end{align}
where $l$ indicates the index of layers in the transformer encoder. The multi-head attention layers have been widely used to model the relationships from channel and spatial dimensions~\cite{vaswani2017attention, dosovitskiy2020image, Hu_2018_CVPR}. Therefore, we argue that it can also be extended to explore the relationships in the batch dimension. As a result, different from typical usage of transformer layers, the input of BatchFormer will be first reshaped to enable the transformer layers working on the batch dimension of the input data. By doing this, the self-attention mechanism in transformer layers then becomes the cross-attention between different samples for BatchFormer.


\textbf{Shared Classifier}. Since we can not assume batch statistics for testing, such as sample relationships, there might be a gap between the features before and after the BatchFormer module. That is, we can not perform inference on new samples by directly removing the BatchFormer. Therefore, apart from the final classifier, we also introduce a new auxiliary classifier to not only learn from the final classifier but also keep consistent with the features before the BatchFormer. To achieves this, we simply share the parameters/weights between the auxiliary classifier and the final classifier. We refer to this simple yet effective strategy as ``shared classifier". With the proposed ``shared classifier", we can thus remove the BatchFormer module during testing, while still benefiting from the sample relationship learning using BatchFormer.

\begin{figure}[!ht]
    \centering
    \includegraphics[width=0.9\linewidth]{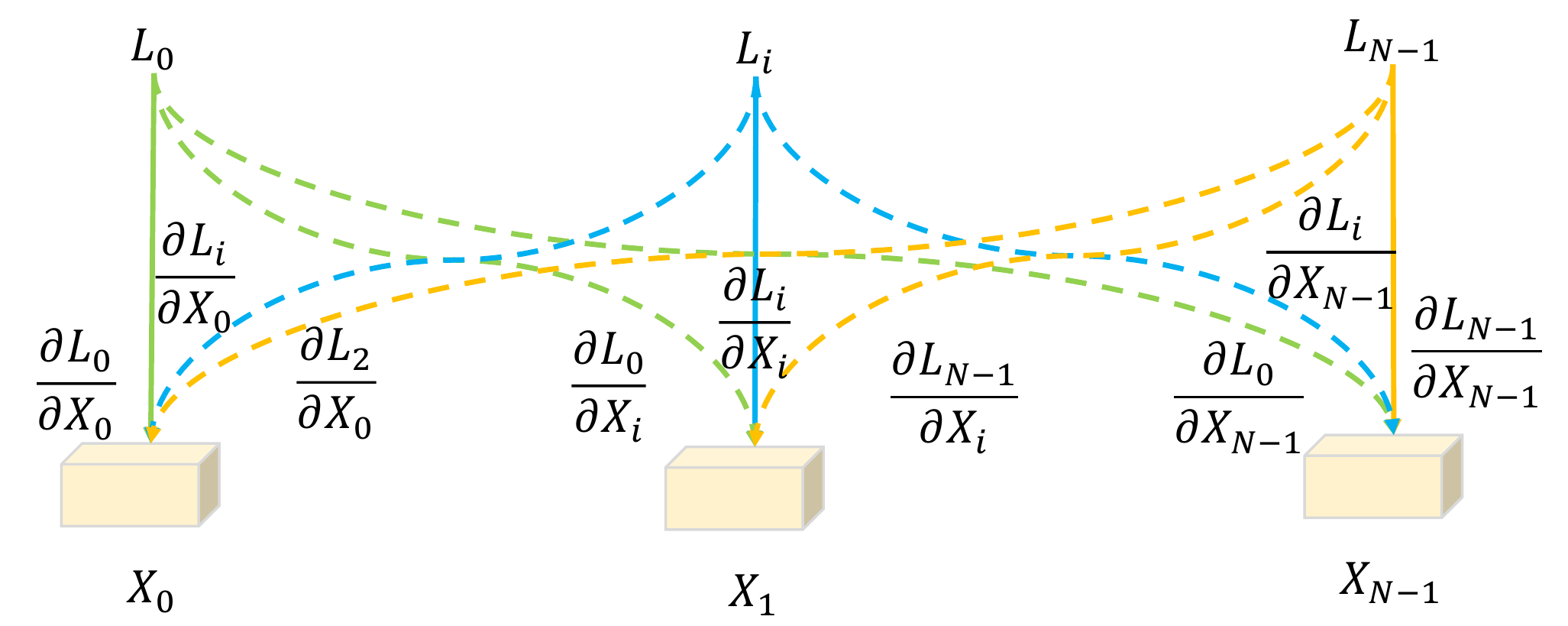}
    \caption{Gradient propagation with BatchFormer in a mini-batch. Dashed lines represent new gradient propagation among samples.}
    \label{fig:grads_illu}
\end{figure}

The proposed BatchFormer is a plug-and-play module for robust deep representation learning. During optimization, the proposed BatchFormer modules can be jointly optimized with other deep learning backbone networks (including both CNNs and Vision Transformers) in an end-to-end way. The proposed BatchFormer is also very easy to implement using typical deep learning softwares,. For example, we show how to implement BatchFormer with several lines of PyTorch code in Algorithm~\ref{alg}.


\subsection{BatchFormer: A Gradient View}





To better understand how the proposed BatchFormer helps representation learning by exploring sample relationships, we also provide an intuitive explanation from the perspective of gradient propagation for optimization. Intuitively, without BatchFormer, all losses only propagate gradients on the corresponding samples and categories, \ie, one-to-one, while there are gradients on other samples with BatchFormer (the dashed line ) as illustrated in Figure~\ref{fig:grads_illu}. Specifically, given samples $X (X=X_0, X_1, X_i,..., X_{N-1})$ and the corresponding losses $L_0, L_1, L_i, ..., L_{N-1}$ in the mini-batch, with BatchFormer, we then have
\begin{equation}
    \label{eq:bt_grads}
    \frac{\partial{L_i}}{\partial{X}} = \frac{\partial{L_i}}{\partial{X_i}} + \sum_{j\neq i}^{N-1}{\frac{\partial{L_i}}{\partial{X_j}}}.
\end{equation}
That is, BatchFormer brings new gradient terms $\frac{\partial{L_i}}{\partial{X_j}}$, where $i \neq j$. From a perspective of gradient optimization, $L_i$ also optimizes the network according to sample $X_j (j\neq i)$, that is a significant difference compared to the model without BatchFormer. In other word, $X_j (j\neq i)$ can be regarded as a virtual sample~\cite{zhang2018mixup, he2021distilling} of $y_i$, where $y_i$ is the label of $X_i$. We consider that both BatchFormer and Mixup~\cite{zhang2018mixup} can be regarded as data-dependent augmentations. BatchFormer implicitly draws virtual examples from the vicinity distribution of samples via cross-attention module. From this perspective, BatchFormer has implicitly augmented $N-1$ virtual samples for each label $y_i$ via the relationship modeling among samples in the mini-batch. Previous approaches~\cite{balaji2018metareg, zhu2018generative} have demonstrated data augmentation is helpful for long-tailed recognition~\cite{he2021distilling, wang2021rsg}, zero-shot learning~\cite{zhu2018generative}, and domain generalization~\cite{zhou2020learning, zhou2021domain}.
the virtual samples are largely helpful for tail classes since those classes lack of samples. Our gradients analysis in Section~\ref{sec:grad_analysis} also demonstrates the tail classes have larger gradients on other samples compared to head classes.

\section{Experiments}

In this section, we perform extensive experiments to demonstrate the effectiveness of BatchFormer for a variety of data scarcity learning tasks, including long-tailed recognition, zero-shot learning, domain generalization and contrastive learning. Appendix provides more results of other tasks, \eg, image classification (include Visual Transformer~\cite{touvron21adeit, dosovitskiy2020image}), and more ablation studies.


\subsection{Long-Tailed Recognition}


{\bf Datasets}. We use four popular datasets for long-tailed recognition as follows: 1) CIFAR-100-LT has 50,000 training images and 10,000 validation images with 100 categories.
2) ImageNet-LT~\cite{liu2019openlongtailrecognition} contains 115.8K images of 1000 classes from ImageNet 2012. The number of images in each class ranges from 5 to 1,280; 3) iNaturalist 2018~\cite{van2018inaturalist} is a large-scale fine-grained dataset with 437.5K images from 8,142 categories; 4) Places-LT~\cite{liu2019openlongtailrecognition} is a long-tailed scene classification dataset derived from the Places dataset~\cite{zhou2017places} with 184.5K images from 365 categories with class instances ranging from 5 to 4,980.

{\bf Baselines}.
We use the following three baseline methods: Balanced Softmax~\cite{ren2020balanced}, RIDE~\cite{wang2020long} and PaCo~\cite{cui2021parametric}. If not otherwise stated, we follow the same settings used in previous methods, and the proposed BatchFormer is removed during testing. Particularly, there is a small difference when comparing with RIDE~\cite{wang2020long}. we train the network with a batch size of 400 on 4 V100 GPUs for 100 epochs with an initial learning rate of 0.1 on ImageNet-LT and 0.2 on iNaturalist 2018, respectively. The learning rare is decayed with cosine schedule on iNaturalist 2018. We report the performance of BatchFormer based on 3 experts RIDE.
All experiments are conducted on V100 GPU with PyTorch. See other details in Appendix A.

\begin{table}[!tp]
\setlength\tabcolsep{2.5pt}
\small
\caption{Illustration of imbalance ratio 100 and 200 on CIFAR-100-LT. $\ast$ means we train the method with the released code in one stage (\eg, balanced softmax~\cite{ren2020balanced}). Med means Medium category. RIDE* means we use 3 experts.}
\label{table:cifar100_lt}
\centering

\begin{tabular}{@{}lcccc|cccc@{}}
\hline
 \multirow{2}{*}{Method} &
\multicolumn{4}{c}{100}&\multicolumn{4}{c}{200} \cr
 & All & Many & Med & Few  & All & Many & Med & Few \\
\hline

RIDE* ~\cite{wang2020long}& 48.0 & 68.1 & 49.2 & 23.9 & - & - & - & -\\
\hline
Balanced$\ast$ ~\cite{ren2020balanced} & 50.7 & 68.0 & 49.7 & 31.9 & 46.4 & 70.0     & 51.5   &  24.3  \\
+ BatchFormer & \bf{51.7} & \bf{68.4} & 49.3 & \bf{34.3} & {\bf 47.5} & {\bf 70.2}  &  {\bf 53.3}    & {\bf 25.5} \\
\hline
Paco~\cite{cui2021parametric} & 51.9 & 63.9 & {\bf 53.0} & {\bf 36.5} & 47.1 & 68.1 & 51.5 & 27.5\\
+ BatchFormer & {\bf 52.4} & {\bf 68.4} & 52.1 & 34.0 & {\bf 47.8} & 68.1 & {\bf 53.0} & {\bf 28.2} \\


\hline
\end{tabular}
\end{table}

\begin{table}[tp]
\small
\setlength\tabcolsep{2.5pt}
\caption{Illustration of ResNet-10 and ResNet-50 on ImageNet-LT. $\ast$ means we train the net work with the released code by one-stage. RIDE-3e means we use three experts in RIDE.}
\label{table:imagenet_lt}
\centering

\begin{tabular}{@{}lcccc|cccc@{}}
\hline
 \multirow{2}{*}{Method} &
\multicolumn{4}{c}{ResNet-10}&\multicolumn{4}{c}{ResNet-50} \cr
 & All & Many & Med & Few  & All & Many & Med & Few \\
\hline
OLTR \cite{liu2019openlongtailrecognition} & 35.6 & 43.2 & 35.1 & 18.5 & - & - & - & \\
LFME~\cite{xiang2020learning} & 38.8 & 47.0 & 37.9 & 19.2 & - & - & - & -\\

\hline
Balanced$\ast$ ~\cite{ren2020balanced} & 41.0 & 52.6 &   38.3 &  18.0 &  50.1 & 61.1  &   47.5   & 27.6    \\
+ BatchFormer\dag & {\bf 43.2} & {\bf 52.8}   &  {\bf 40.7}  &  {\bf 24.9}  & {\bf 51.1} & {\bf 61.4}   &  {\bf 47.8}  &  {\bf 33.6}       \\
\hline
RIDE-3e $\ast$ ~\cite{wang2020long} & 44.7 & 57.0 & 40.3 & 25.5 & 53.6 & 64.9 & 50.4 & 33.2 \\
+ BatchFormer & {\bf 45.7} & 56.3 & {\bf 42.1} & {\bf 28.3} & {\bf 54.1} & 64.3 & {\bf 51.4} & {\bf 35.1}\\
\hline
PaCo~\cite{cui2021parametric} & - & - & - & - & 57.0 & 64.8 & 55.9 & 39.1 \\
+ BatchFormer & - & - & - & - & {\bf 57.4} & 62.7 & {\bf 56.7} & {\bf 42.1}\\
\hline\hline
two stage \\
\hline
RIDE-3e ~\cite{wang2020long} & 45.9 & 57.6 & 41.7 & 28.0 & 54.9 & 66.2 & 51.7 & 34.9 \\
+ BatchFormer & {\bf 47.6} &  55.3 & {\bf 45.5} & {\bf 33.3} & {\bf 55.7} & 64.6 & {\bf 53.4} & {\bf 39.0} \\
\hline
\end{tabular}
\end{table}




{\bf Results on CIFAR-100-LT}. Table~\ref{table:cifar100_lt} demonstrates the proposed BatchFormer module is orthogonal to the state-of-the-art methods, \eg, Balanced Softmax and Paco. We notice that BatchFormer largely improves Balanced Softmax by 2.4\% for few classes when imbalance ratio is 100, and by 1.8\% on medium classes and by 1.2\% on few classes respectively when imbalance ratio is 200. Besides, the results of PaCo on imbalance ratio 200 also increase by 1.5\% on medium classes and 0.7\% on few classes with BatchFormer. For ratio 100, BatchFormer mainly improves many classes since Paco has achieved good performance on few classes. Please refer to Appendix for more explanations (\ie, the comparison without PaCo loss, but with strong data augmentation). Overall, BatchFormer improves the recognition of tail classes while maintaining the performance of head classes. Meanwhile BatchFormer is pluggable to current popular methods on CIFAR-100-LT.

{\bf  Results on ImageNet-LT}. As illustrated in Table~\ref{table:imagenet_lt}, BatchFormer largely improves Balanced Softmax by 2.4\% on medium classes and {\bf 6.9\%} on few classes respectively under ResNet-10 backbone. Meanwhile, under ResNet-50 backbone, BatchFormer largely improves Balanced Softmax on the Few category by {\bf 5\%}.

When BatchFormer is applied in RIDE~\cite{wang2020long}, the results on medium classes and few classes increase by 1.8\% and 2.8\% respectively under ResNet-10 backbone. BatchFormer also increases medium and few classes by 1\% and 1.9\% respectively under ResNet-50. BatchFormer achieves clear improvement overall, while the performance on many classes drops a bit. Furthermore, BatchFormer also effectively improves RIDE under two-stage training strategy (RIDE uses a larger model to teach small model with distilling loss). Here, different from RIDE, we use a pre-trained model (the same model) to initialize the model and train the model again with BatchFormer (See details in Appendix).

PaCo~\cite{cui2021parametric} is recently introduced for Long-Tailed Recognition with Supervised Contrastive Learning. We also find BatchFormer is able to facilitate ImageNet-LT on medium classes and few classes. Noticeably, PaCo uses a strong data augmentation strategy from Supervised Contrastive Learning with 400 training epochs. We think data augmentation limits the improvement of BatchFormer on ImageNet-LT. Particularly, Appendix also shows BatchFormer achieves comparable results on balanced Full ImageNet.

\begin{table}[tp]
\small
\caption{Illustration on iNaturalist 2018. $\ast$ means we train the method with the released code in one stage. RIDE-3e means RIDE with 3 experts.}
\label{table:inaturalist}
\centering

\begin{tabular}{@{}lcccc@{}}
\hline
Methods & All & Many & Medium & Few \\
\hline
BBN~\cite{zhou2020bbn} & 66.3 & 49.4 & 70.8 & 65.3 \\
cRT~\cite{kang2019decoupling} & 65.2 & 69.0 & 66.0 & 63.2 \\
RIDE-3e~\cite{wang2020long} & 72.2 & 70.2 & 72.2 & 72.7 \\
PaCo~\cite{cui2021parametric} & 73.2 & - & - & -\\
\hline
RIDE-3e$\ast$~\cite{wang2020long}  & 72.5 & 68.1 & 72.7 & 73.2 \\
+ Batchformer & {\bf 74.1} & 65.5 & {\bf 74.5} & {\bf 75.8} \\


\hline
\end{tabular}
\end{table}

{\bf  Results on iNaturalist 2018}. We mainly evaluate BatchFormer on RIDE since Balanced Softmax has limited performance on iNaturalist 2018 and PaCo requires over 36 GPU days to converge (See Appendix). We train RIDE with cosine decay learning-rate scheduler by one-stage and achieve a strong baseline with 72.5\% (better than the reported) as illustrated in Table~\ref{table:inaturalist}. Particularly, BatchFormer further improves the Medium category and Few Category by 1.8\% and 2.6\% respectively. Here, for a fair comparison to previous work, we mainly evaluate BatchFormer RIDE with 3 experts which has similar GFlops to previous work.

\begin{table}[tp]
\small
\caption{Illustration of Places-LT (Backbone is ResNet-152).  $\ast$ means the result that we train the method with the released code.}
\label{table:places_lt}
\centering

\begin{tabular}{@{}lcccc@{}}
\hline
Methods & All & Many & Medium & Few \\
\hline

OLTR~\cite{liu2019openlongtailrecognition} & 35.9 & 44.7 & 37.0 & 25.3 \\
$\tau$-normalized~\cite{kang2019decoupling} & 37.9 & 37.8 & 40.7 & 31.8 \\
\hline
BALMS*~\cite{ren2020balanced} & 37.8 & 41.4 &	   38.8 &	 29.1 \\
+ Batchformer & {\bf 38.2} & 39.5  &   38.3  &  {\bf 35.7} \\
\hline
PaCo~\cite{cui2021parametric} &  41.2 & 37.5 & {\bf 47.2} & 33.9 \\
+ Batchformer &  {\bf 41.6} & {\bf 44.0} & 43.1 & 33.7 \\
\hline
\end{tabular}
\end{table}

{\bf  Results on Places-LT}. Table~\ref{table:places_lt} illustrates BatchFormer improves BALMS on Few category. Besides, BatchFormer effectively improves PaCo~\cite{cui2021parametric} which is the new state-of-the-art on Places-LT. Here, different from CIFAR-100-LT and ImageNet-LT, BatchFormer mainly improves Many category. This might be because the result of PaCo on Many category is very worse and BatchFormer can re-balance the imbalanced datasets.

\subsection{Zero-Shot Learning}
We evaluate BatchFormer on compositional zero-shot learning~\cite{naeem2021learning}. All experiments are evaluated with one V100 GPU. The learning rare, epochs, optimizer are fully similar to ~\cite{naeem2021learning}. See details in Appendix A.

{\bf Datasets.} The experiments are performed on three datasets: 1) MIT-States~\cite{isola2015discovering} consists of 30,000 training images of natural objects with 1,262 seen compositions (23.8 image per composition, 115 states and 245 objects), and 13,000 test images with 400 seen compositions and 400 unseen compositions; 2) UT-Zappos~\cite{yu2014fine} includes 23,000 training images of shoes catalogue and we use the splits from~\cite{purushwalkam2019task}. UT-Zappos has 83 seen compositions for training (277.1 images per composition, 16 states, 12 objects) and 18 seen compositions and unseen compositions in test set; 3) C-GQA~\cite{naeem2021learning} provides 26,000 training images with 6,963 seen compositions (3.7 images per composition, 453 states, 870 objects) and 3,000 test images with 18 seen compositions and unseen compositions.

{\bf Metrics.} We adopt the evaluation protocol of ~\cite{purushwalkam2019task} and report the Area Under the Curve (AUC) (in \%) between the accuracy on seen and unseen compositions. Similar to~\cite{naeem2021learning}, we also report unseen accuracy and seen accuracy, as well as the best harmonic mean.

Table~\ref{table:illu_zs} demonstrates BatchFormer effectively improves the AUC and HM among all the datasets compared to the baseline. Particularly, for a fair comparison, we use the released code under the same setting to reproduce ~\cite{naeem2021learning} as the baseline. We notice BatchFormer mainly improves the Seen category on MIT-States and C-GQA, while BatchFormer largely improves the unseen category by nearly {\bf 5\%}. This might be because the number of seen composition instances on MIT-States and C-GQA is few, \eg 23.8 image per seen composition on MIT-States and 3.7 images per seen composition. In other words, the recognition of seen compositions on MIT-States and C-GQA is few-shot learning. We think BatchFormer on the two datasets mainly finds invariant features among images of the same class.

\begin{table*}[tp]
\setlength\tabcolsep{4pt}
\small
\caption{Illustration of Batchformer on Compositional Zero-Shot Learning. * means we use the released code to reproduce the results. S means seen, U means unseen, s means state and o means object. The results of ~\cite{li2020symmetry} are copied from ~\cite{naeem2021learning}. }
\label{table:illu_zs}
\centering

\begin{tabular}{@{}lcccccc|cccccc|cccccc@{}}
\hline
 \multirow{2}{*}{Method} &
\multicolumn{6}{c}{MIT-States}&\multicolumn{6}{c}{UT-Zap50K} &\multicolumn{6}{c}{C-GQA} \cr
 &  AUC & HM & S & U & s & o & AUC & HM & S & U  & s & o & AUC & HM & S & U  & s & o\\
\hline
CompCos~\cite{mancini2021open} & 4.5 & 16.4 & 25.3 & 24.6 & 27.9 & 31.8 & 28.7 & 43.1 & 59.8 & 62.5&44.7 & 73.5&  - & - & -& -& -& -\\

CGE~\cite{naeem2021learning} & 6.5 & 21.4 & 32.8 & 28.0& 30.1& 34.7& 33.5 &60.5& 64.5& 71.5& 48.7& 76.2 &3.6& 14.5& 31.4& 14.0& 15.2& 30.4 \\
\hline
CGE* & 6.3 & 20.0 & 31.6 & 27.3 & 30.3 & 34.5 & 31.5 & 46.5 & 60.3 & 64.5 & 46.3 & 74.4 & 3.7 & 14.9 & 30.8 & 14.7 & {\bf 15.8} & 29.0 \\
+BatchFormer & {\bf 6.7} & {\bf 20.6} & {\bf 33.2} & {\bf 27.7} & {\bf 30.8} & {\bf 34.7} & {\bf 34.6} & {\bf 49.0}  & {\bf 62.5 }& {\bf 69.2 }& {\bf 49.7} & {\bf 75.6} & {\bf 3.8} & {\bf 15.5} & {\bf 31.3} & 14.7 & 15.3 & {\bf 30.0} \\


\hline
\end{tabular}
\end{table*}

\subsection{Domain Generalization}

By exploring sample relationships of the same class, it is easier to find the invariant features and thus improve the domain generalization. We first demonstrate BatchFormer effectively improves the baseline (without domain generalization techniques) on PACS~\cite{li2017deeper} under ResNet-18. Then, we apply BatchFormer to recent domain generalization, \eg, SWAD~\cite{cha2021swad}, and shows the effectiveness of BatchFormer.

{\bf Datasets}. We illustrates the application of BatchFormer on several domain generalization datasets: 1) PACS~\cite{li2017deeper} covers 7 object categories and 4 domains (Photo, Art Paintings, Cartoon and Sketches), 2) VLCS~\cite{fang2013unbiased} (4 domains, 5 classes, 10,729 images), OfficeHome~\cite{venkateswara2017deep} (4 domains, 65 classes and 15,588 images), and TerraIncognita~\cite{beery2018recognition} (4 domains, 10 classes and 24,788 images).

{\bf Details}. For baseline, we train the network under ResNet-18 with SGD (30 epochs, initial learning rate 0.001), and drop the learning rate at 24 epochs. We also use the popular data augmentations, \eg, flip, color jiter and scale. We train the network 5 times and report the average. Others methods are compared based on ~\cite{dalib}. For a fair comparison with SWAD~\cite{cha2021swad}, we follow the optimization methods of~\cite{cha2021swad} and use the released code of ~\cite{cha2021swad} to evaluate BatchFormer on ResNet-18. See more details in Appendix.

Table~\ref{table:dg_pacs} illustrates BatchFormer consistently improves baseline , CORAL~\cite{deep_coral} and MixStyle~\cite{zhou2021mixstyle}. BatchFormer also clearly improves recent work~\cite{cha2021swad} on four datasets in Table~\ref{table:dg_swad}. Particularly, BatchFormer improves~\cite{cha2021swad} by over 2\% on OfficeHome and TerraIncognita. This illustrates BatchFormer is able to facilitate invariant representation learning, and improve the generalization across domains.

\begin{table}[tp]
\setlength\tabcolsep{2.5pt}
\small
\caption{Illustration of BatchFormer for Domain Generalization under different works on PACS~\cite{li2017deeper}.}
\label{table:dg_pacs}
\centering
\begin{tabular}{@{}lccccc@{}}
\hline
Methods  & art\_paint & cartoon & sketches & photo & Avg.  \\
\hline
Baseline &79.9$\pm$ 1.0 & 73.0$\pm$1.5 & 67.7$\pm$ 3.0& 95.7$\pm$0.4 & 79.1\\
+BatchFormer & {\bf 80.4}$\pm$0.2 & {\bf 73.8}$\pm$2.0& {\bf 68.6}$\pm$1.8&{\bf 96.3}$\pm$0.2 & {\bf 79.8}\\
\hline
CORAL~\cite{deep_coral} & 79.2$\pm$1.7 &  {\bf 75.5} $\pm$1.1 & 71.4$\pm$3.1 & 94.7$\pm$0.3 &  80.2 \\
+BatchFormer & {\bf 80.6}$\pm$0.9 &  74.7$\pm$1.9 & {\bf 73.1}$\pm$0.3& {\bf 95.1}$\pm$0.3 & {\bf 80.9} \\
\hline
MixStyle~\cite{zhou2021mixstyle} & 81.7$\pm$0.1 &  {\bf 76.8}$\pm$0.0 & 80.8$\pm$0.0 &  93.1$\pm$0.0 & 83.1 \\
+BatchFormer & {\bf 84.8} $\pm$0.4 &  75.3$\pm$0.0 &  {\bf 81.1} $\pm$0.4 & {\bf 93.6}$\pm$0.0 &  {\bf 83.7}\\


\hline
\end{tabular}
\end{table}

\begin{table}[tp]
\small
\caption{Illustration of BatchFormer for Domain Generalization based on recent work~\cite{cha2021swad} (ResNet-18). We show the average results of different domains. Terra is TerraIncognita. See more results in Appendix. }
\label{table:dg_swad}
\centering
\begin{tabular}{@{}lcccc@{}}
\hline
Methods  & PACS & VLCS & OfficeHome & Terra   \\
\hline
SWAD*~\cite{cha2021swad} & 82.9 & 76.3& 62.1 & 42.1 \\
 + BatchFormer & {\bf 83.7} & {\bf 76.9} & {\bf 64.3} & {\bf 44.8}\\
\hline
\end{tabular}
\end{table}


\begin{table}[tp]
\small
\caption{Illustration of BatchFormer for Contrastive Learning (MoCo~\cite{he2019moco, chen2020mocov2, chen2021mocov3} under ResNet50) on linear  classification.}
\label{table:moco}
\centering
\begin{tabular}{@{}lccc@{}}
\hline
Methods & Epochs & Top-1 & Top-5 \\
\hline
MoCo-v2~\cite{chen2020mocov2}    & 200 & 67.5 & -\\
MoCo-v2~\cite{chen2020mocov2}+BatchFormer & 200 & {\bf 68.4} & 88.5\\
\hline
MoCo-v3~\cite{chen2021mocov3} & 100 & 68.9 & - \\
MoCo-v3~\cite{chen2021mocov3}+BatchFormer & 100 & {\bf 69.8} & 89.5 \\
\hline
\end{tabular}
\end{table}

\begin{figure}
    \centering
    \includegraphics[width=.4\textwidth]{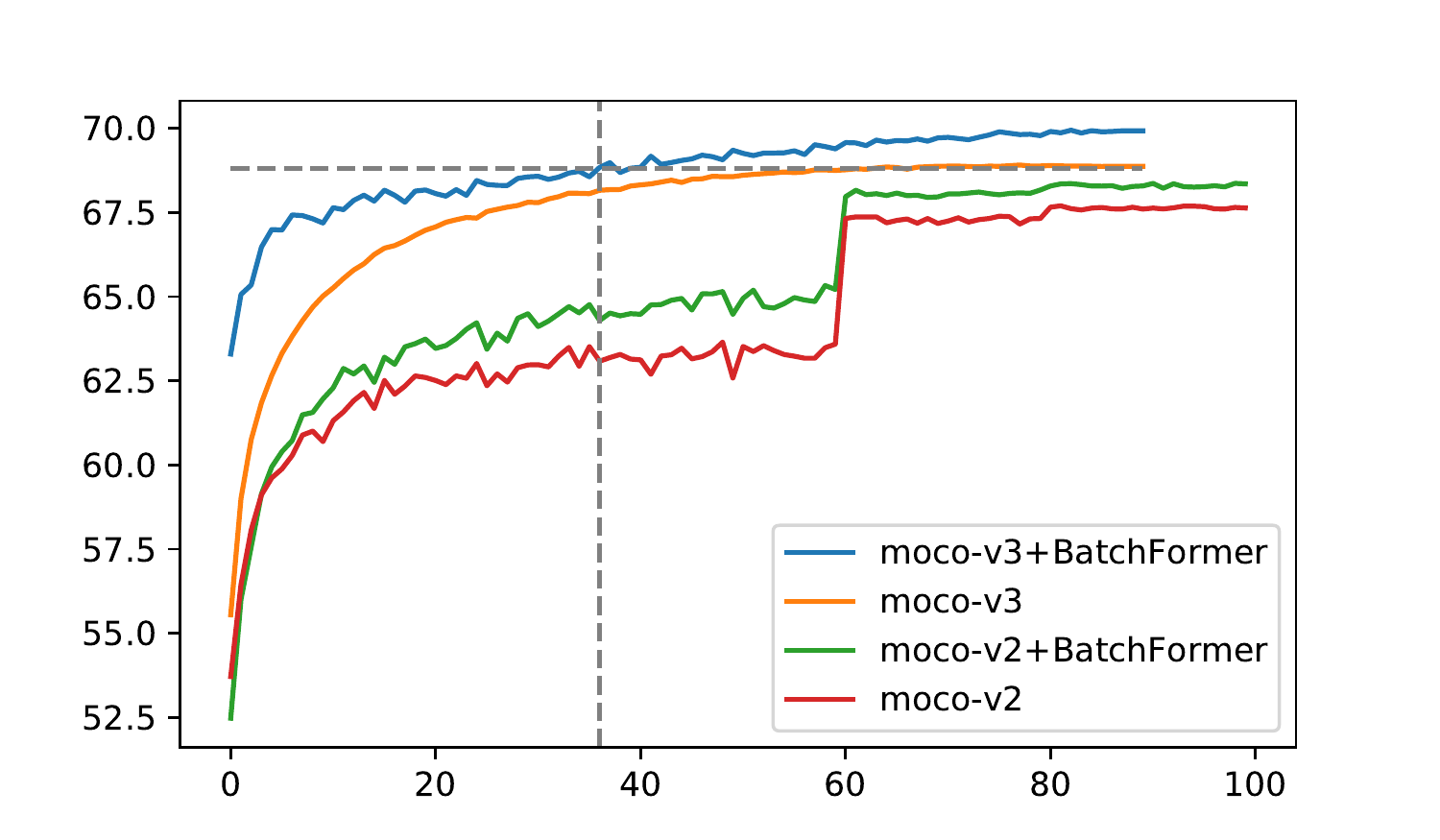}
    \caption{Convergence of BatchFormer under linear classification. The batch size and optimizer are the same as~\cite{chen2021mocov3}.}
    \label{fig:conveg_bt}
\end{figure}

\subsection{Self-Supervised Learning}
Contrastive Learning aims to learn representations that attract similar samples and dispel different samples, while BatchFormer builds a Transformer Network among samples to implicitly explore the sample relationships for representation learning. BatchFormer can also be applied to Contrastive Learning. We mainly evaluate BatchFormer with MoCo-v2~\cite{chen2020mocov2} and MoCo-v3~\cite{chen2021mocov3} on linear classification protocol. Object detection result is provided in Appendix.

Table~\ref{table:moco} shows BatchFormer is also able to improve the representation learning of self-supervised learning, \eg, BatchFormer consistently improves MoCo-v2 and MoCo-v3 by around 1.\% on ImageNet. Figure~\ref{fig:conveg_bt} shows the linear classification of BatchFormer pre-trained model convergences faster, \eg, BatchFormer pre-trained model achieves the performance of moco-v3 with only 38 epochs.


\subsection{Ablation Studies}
\label{sec:ab}


\begin{figure*}
    \centering
    \includegraphics[width=.8\textwidth]{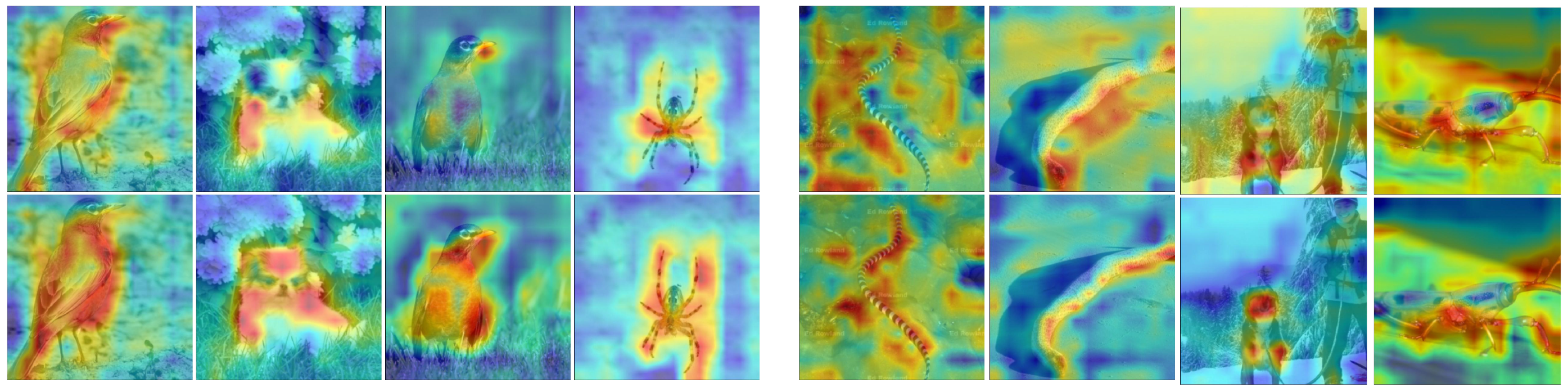}
    \caption{Grad-Cam demonstration of BatchFormer on low-shot test images based on~\cite{ren2020balanced}. The first row is baseline, while the second row is BatchFormer. The left images show BatchFormer enables the model pay attention on more details when the scene is simple and clean, while the right images show BatchFormer faciliates the model ignore the spurious correlation in the image. More figures are in Appendix.}
    \label{fig:cam_illu}
\end{figure*}

\begin{figure}
    \centering
    \includegraphics[width=.4\textwidth]{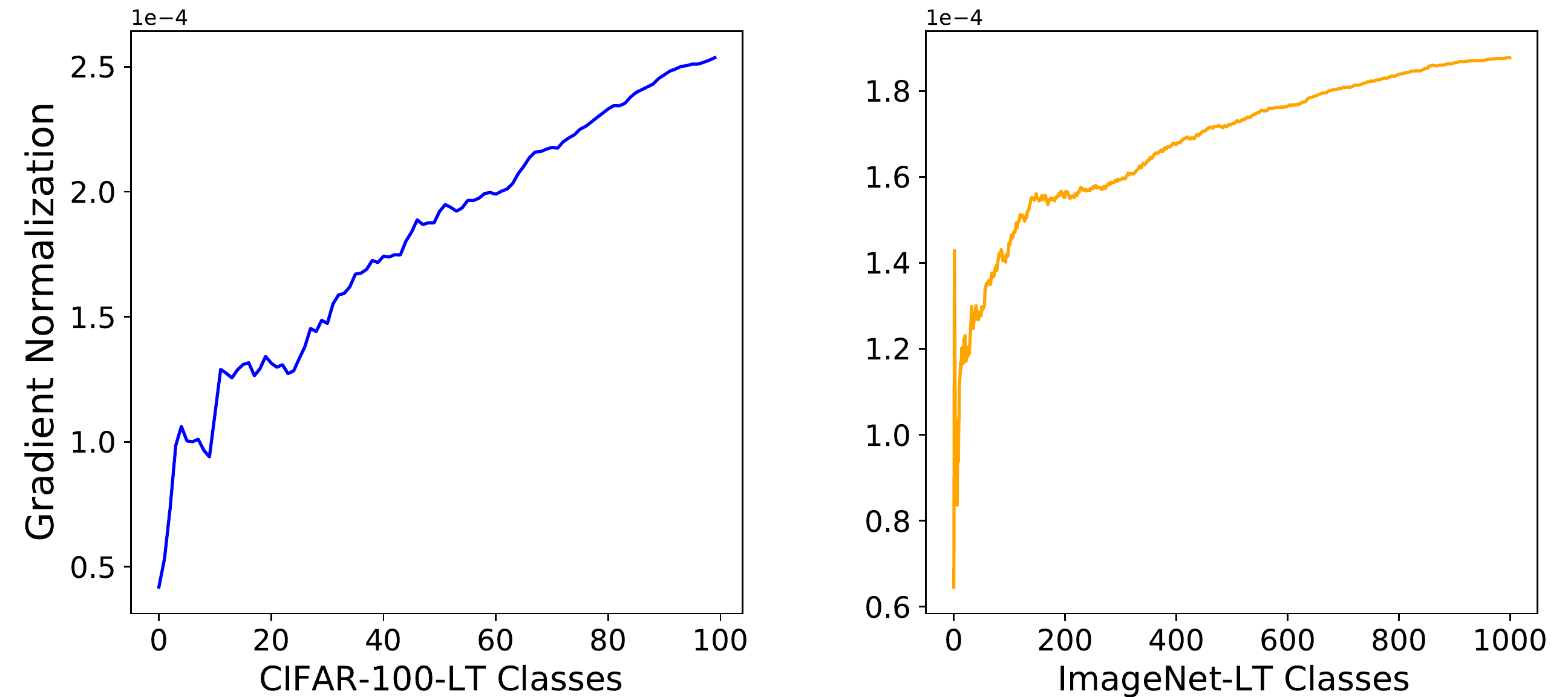}
    \caption{The gradient of each class to other images in mini-batch on CIFAR-100-LT and ImageNet-LT (based on~\cite{ren2020balanced}). For each class, we obtain the gradient normalization of the loss of each label to other images in the mini-batches, and then average the gradients of each class in the test set. The classes are sorted by descending order according to the number of instances in the training set.}
    \label{fig:grad}
\end{figure}



{\bf Batch size.} The batch size represents the size of Transformer. Table~\ref{table:ab_batch_size} shows BatchFormer is less sensitive to the batch size when batch size is less than 128. We find batch size 512 achieves better performance on Few category.

\begin{table}[tp]
\small
\caption{Ablation Studies of Batch Size based on Balanced Softmax~\cite{ren2020balanced}. The backbone is ResNet-10 under one GPU. learning rare is linearly decreased with batch size.}
\label{table:ab_batch_size}
\centering
\begin{tabular}{@{}lcccccc@{}}
\hline
Batch Size & 16 & 32 & 64 & 128 & 256 & 512  \\

\hline
All &  43.2  & {\bf 43.6} & 43.3 & 43.2 & 42.4 &42.5\\
Many & {\bf 53.8} & 53.7 & 52.9&   52.8 & 52.2 & 52.0\\
Medium & 40.2 & 40.7 & {\bf 40.8} &   40.7 &    39.5 & 39.6\\
Few & 23.9  &24.9 &  25.3 &  24.9 &  25.2 & {\bf 25.8 }\\
\hline
\end{tabular}
\end{table}

\begin{table}[tp]
\small
\caption{Ablation Studies on ImageNet-LT based on Balanced Softmax~\cite{ren2020balanced}. The backbone is ResNet-50.}
\label{table:ab_modules}
\centering

\begin{tabular}{@{}lcccc@{}}
\hline
Method & All & Many & Medium & Few \\
\hline
BatchFormer & {\bf 50.9} & {\bf 60.7} & {\bf 47.7} & 34.1 \\
w/o Shared Classifier &42.4 & 41.3  & 43.3 & {\bf 42.2} \\
\hline
\end{tabular}
\end{table}


{\bf Shared Classifier}. Interestingly, Table~\ref{table:ab_modules} demonstrates BatchFormer without shared classifier achieves similar performance on three categories, while BatchFormer with shared classifier maintains the performance on Many category. This demonstrates the effectiveness of shared classifier and the ability of re-balancing of BatchFormer.

\section{Visualization Analysis}
\label{sec:visualize}
\label{sec:grad_analysis}

{\bf Visualized Comparison}.
We illustrate the visualized comparison with Grad-CAM~\cite{selvaraju2017grad} between Baseline and BatchFormer in Figure~\ref{fig:cam_illu}. We find BatchFormer focuses on more details of objects and ignores spurious correlations.
On the one hand, when the image includes complex scence with many disturbing factors, BatchFormer effectively improves the attention of the network on the corresponding object regions (\eg, sea snake on the sandbeach, dog on the snow and insect on the leaf in Figure~\ref{fig:cam_illu} right). On the other hand, BatchFormer also pays more attention on regions of the object when the scene is clear (\eg, the bird, dog, and spider in Figure~\ref{fig:cam_illu} left). See more examples in Appendix.

{\bf Gradients Analysis}.
BatchFormer has increased new gradient backward: the loss of each label has the gradients on other images. In other words, we have implicitly augmented the samples for the class of each image in the mini-batches. The other images in the mini-batch can also be regarded as the virtual instances of current image class.  The gradient is firmly related to the effect of each image label on other images. Figure~\ref{fig:grad} illustrates the rarer the class is, the larger the gradients of the class have on other images in the mini-batch. Thus, BatchFormer actually utilizes the other images to facilitate low-shot recognition via increasing gradients of few-shot labels on other images.


\section{Conclusion and Future work}

We propose to enable deep neural networks themselves with the ability to explore the sample relationships from each mini-batch. Specifically, we regards each image (batch  dimension) in the mini-batch as a node of a sequence, and then build a Transformer Encoder Network among the images to mine the relationships among the images in the mini-batch. BatchFormer enables the gradient propagation of each label to all images in the mini-batch, which can be regarded as virtual sample augmentation, and thus improve the representation learning. We further introduce a shared classifier before and after the BatchFormer during training,  which can thus be removed during testing.  We demonstrate the effectiveness of BatchFormer on over ten datasets and BatchFormer achieves significant improvements on different tasks, including long-tailed recognition, zero-shot learning, domain generalization, and contrastive learning. \\
{\small
{\bf Limitations} The improvement of current BatchFormer on models with strong data augmentation and balanced distribution is limited. \\
{\bf Acknowledgements} {\small Dr. Baosheng Yu and Mr. Zhi Hou are supported by ARC FL-170100117.}}

{\small
\bibliographystyle{ieee_fullname}
\bibliography{egbib}

\begin{thebibliography}{10}\itemsep=-1pt

\bibitem{arjovsky2019invariant}
Martin Arjovsky, L{\'e}on Bottou, Ishaan Gulrajani, and David Lopez-Paz.
\newblock Invariant risk minimization.
\newblock {\em arXiv preprint arXiv:1907.02893}, 2019.

\bibitem{atzmon2020causal}
Yuval Atzmon, Felix Kreuk, Uri Shalit, and Gal Chechik.
\newblock A causal view of compositional zero-shot recognition.
\newblock In {\em NeurIPS}, 2020.

\bibitem{balaji2018metareg}
Yogesh Balaji, Swami Sankaranarayanan, and Rama Chellappa.
\newblock Metareg: Towards domain generalization using meta-regularization.
\newblock In {\em NIPS}, volume~31, pages 998--1008, 2018.

\bibitem{beery2018recognition}
Sara Beery, Grant Van~Horn, and Pietro Perona.
\newblock Recognition in terra incognita.
\newblock In {\em ECCV}, pages 456--473, 2018.

\bibitem{blanchard2011generalizing}
Gilles Blanchard, Gyemin Lee, and Clayton Scott.
\newblock Generalizing from several related classification tasks to a new
  unlabeled sample.
\newblock {\em NIPS}, 24:2178--2186, 2011.

\bibitem{byrd2019effect}
Jonathon Byrd and Zachary Lipton.
\newblock What is the effect of importance weighting in deep learning?
\newblock In {\em ICML}, pages 872--881. PMLR, 2019.

\bibitem{cai2021ace}
Jiarui Cai, Yizhou Wang, and Jenq-Neng Hwang.
\newblock Ace: Ally complementary experts for solving long-tailed recognition
  in one-shot.
\newblock In {\em ICCV}, 2021.

\bibitem{cha2021swad}
Junbum Cha, Sanghyuk Chun, Kyungjae Lee, Han-Cheol Cho, Seunghyun Park, Yunsung
  Lee, and Sungrae Park.
\newblock Swad: Domain generalization by seeking flat minima.
\newblock In {\em NeurIPS}, 2021.

\bibitem{chattopadhyay2020learning}
Prithvijit Chattopadhyay, Yogesh Balaji, and Judy Hoffman.
\newblock Learning to balance specificity and invariance for in and out of
  domain generalization.
\newblock In {\em ECCV}, pages 301--318. Springer, 2020.

\bibitem{chawla2002smote}
Nitesh~V Chawla, Kevin~W Bowyer, Lawrence~O Hall, and W~Philip Kegelmeyer.
\newblock Smote: synthetic minority over-sampling technique.
\newblock {\em Journal of artificial intelligence research}, 16:321--357, 2002.

\bibitem{chen2020simple}
Ting Chen, Simon Kornblith, Mohammad Norouzi, and Geoffrey Hinton.
\newblock A simple framework for contrastive learning of visual
  representations.
\newblock In {\em ICML}, pages 1597--1607. PMLR, 2020.

\bibitem{chen2020mocov2}
Xinlei Chen, Haoqi Fan, Ross Girshick, and Kaiming He.
\newblock Improved baselines with momentum contrastive learning.
\newblock {\em arXiv preprint arXiv:2003.04297}, 2020.

\bibitem{chen2021mocov3}
Xinlei Chen*, Saining Xie*, and Kaiming He.
\newblock An empirical study of training self-supervised vision transformers.
\newblock In {\em CVPR}, 2021.

\bibitem{cui2021parametric}
Jiequan Cui, Zhisheng Zhong, Shu Liu, Bei Yu, and Jiaya Jia.
\newblock Parametric contrastive learning.
\newblock In {\em ICCV}, 2021.

\bibitem{cui2019class}
Yin Cui, Menglin Jia, Tsung-Yi Lin, Yang Song, and Serge Belongie.
\newblock Class-balanced loss based on effective number of samples.
\newblock In {\em CVPR}, pages 9268--9277, 2019.

\bibitem{dosovitskiy2020image}
Alexey Dosovitskiy, Lucas Beyer, Alexander Kolesnikov, Dirk Weissenborn,
  Xiaohua Zhai, Thomas Unterthiner, Mostafa Dehghani, Matthias Minderer, Georg
  Heigold, Sylvain Gelly, et~al.
\newblock An image is worth 16x16 words: Transformers for image recognition at
  scale.
\newblock In {\em ICLR}, 2020.

\bibitem{everingham2010pascal}
Mark Everingham, Luc Van~Gool, Christopher~KI Williams, John Winn, and Andrew
  Zisserman.
\newblock The pascal visual object classes (voc) challenge.
\newblock {\em International journal of computer vision}, 88(2):303--338, 2010.

\bibitem{fang2013unbiased}
Chen Fang, Ye Xu, and Daniel~N Rockmore.
\newblock Unbiased metric learning: On the utilization of multiple datasets and
  web images for softening bias.
\newblock In {\em ICCV}, pages 1657--1664, 2013.

\bibitem{ghiasi2021simple}
Golnaz Ghiasi, Yin Cui, Aravind Srinivas, Rui Qian, Tsung-Yi Lin, Ekin~D Cubuk,
  Quoc~V Le, and Barret Zoph.
\newblock Simple copy-paste is a strong data augmentation method for instance
  segmentation.
\newblock In {\em CVPR}, pages 2918--2928, 2021.

\bibitem{hadsell2006dimensionality}
Raia Hadsell, Sumit Chopra, and Yann LeCun.
\newblock Dimensionality reduction by learning an invariant mapping.
\newblock In {\em CVPR}, volume~2, pages 1735--1742. IEEE, 2006.

\bibitem{han2005borderline}
Hui Han, Wen-Yuan Wang, and Bing-Huan Mao.
\newblock Borderline-smote: a new over-sampling method in imbalanced data sets
  learning.
\newblock In {\em International conference on intelligent computing}, pages
  878--887. Springer, 2005.

\bibitem{he2009learning}
Haibo He and Edwardo~A Garcia.
\newblock Learning from imbalanced data.
\newblock {\em IEEE Transactions on Knowledge and Data Engineering},
  21(9):1263--1284, 2009.

\bibitem{he2021compas}
Ju He, Adam Kortylewski, and Alan Yuille.
\newblock Compas: Representation learning with compositional part sharing for
  few-shot classification.
\newblock {\em arXiv preprint arXiv:2101.11878}, 2021.

\bibitem{he2019moco}
Kaiming He, Haoqi Fan, Yuxin Wu, Saining Xie, and Ross Girshick.
\newblock Momentum contrast for unsupervised visual representation learning.
\newblock In {\em CVPR}, 2020.

\bibitem{He2015}
Kaiming He, Xiangyu Zhang, Shaoqing Ren, and Jian Sun.
\newblock Deep residual learning for image recognition.
\newblock In {\em CVPR}, 2016.

\bibitem{he2021distilling}
Yin-Yin He, Jianxin Wu, and Xiu-Shen Wei.
\newblock Distilling virtual examples for long-tailed recognition.
\newblock In {\em ICCV}, 2021.

\bibitem{hjelm2018learning}
R~Devon Hjelm, Alex Fedorov, Samuel Lavoie-Marchildon, Karan Grewal, Phil
  Bachman, Adam Trischler, and Yoshua Bengio.
\newblock Learning deep representations by mutual information estimation and
  maximization.
\newblock In {\em ICLR}, 2019.

\bibitem{hou2019cross}
Ruibing Hou, Hong Chang, Bingpeng Ma, Shiguang Shan, and Xilin Chen.
\newblock Cross attention network for few-shot classification.
\newblock {\em NeurIPS}, 2019.

\bibitem{hou2020visual}
Zhi Hou, Xiaojiang Peng, Yu Qiao, and Dacheng Tao.
\newblock Visual compositional learning for human-object interaction detection.
\newblock In {\em ECCV}, pages 584--600. Springer, 2020.

\bibitem{hou2021fcl}
Zhi Hou, Baosheng Yu, Yu Qiao, Xiaojiang Peng, and Dacheng Tao.
\newblock Detecting human-object interaction via fabricated compositional
  learning.
\newblock In {\em CVPR}, pages 14646--14655, 2021.

\bibitem{Hu_2018_CVPR}
Jie Hu, Li Shen, and Gang Sun.
\newblock Squeeze-and-excitation networks.
\newblock In {\em CVPR}, June 2018.

\bibitem{huang2020self}
Zeyi Huang, Haohan Wang, Eric~P Xing, and Dong Huang.
\newblock Self-challenging improves cross-domain generalization.
\newblock In {\em ECCV}, pages 124--140. Springer, 2020.

\bibitem{ioffe2015batch}
Sergey Ioffe and Christian Szegedy.
\newblock Batch normalization: Accelerating deep network training by reducing
  internal covariate shift.
\newblock In {\em ICML}, pages 448--456. PMLR, 2015.

\bibitem{isola2015discovering}
Phillip Isola, Joseph~J Lim, and Edward~H Adelson.
\newblock Discovering states and transformations in image collections.
\newblock In {\em CVPR}, pages 1383--1391, 2015.

\bibitem{jamal2020rethinking}
Muhammad~Abdullah Jamal, Matthew Brown, Ming-Hsuan Yang, Liqiang Wang, and
  Boqing Gong.
\newblock Rethinking class-balanced methods for long-tailed visual recognition
  from a domain adaptation perspective.
\newblock In {\em CVPR}, pages 7610--7619, 2020.

\bibitem{dalib}
Junguang Jiang, Baixu Chen, Bo Fu, and Mingsheng Long.
\newblock Transfer-learning-library.
\newblock \url{https://github.com/thuml/Transfer-Learning-Library}, 2020.

\bibitem{kang2019decoupling}
Bingyi Kang, Saining Xie, Marcus Rohrbach, Zhicheng Yan, Albert Gordo, Jiashi
  Feng, and Yannis Kalantidis.
\newblock Decoupling representation and classifier for long-tailed recognition.
\newblock {\em ICLR}, 2020.

\bibitem{kato2018compositional}
Keizo Kato, Yin Li, and Abhinav Gupta.
\newblock Compositional learning for human object interaction.
\newblock In {\em ECCV}, pages 234--251, 2018.

\bibitem{VREx}
David Krueger, Ethan Caballero, Joern-Henrik Jacobsen, Amy Zhang, Jonathan
  Binas, Dinghuai Zhang, Remi~Le Priol, and Aaron Courville.
\newblock Out-of-distribution generalization via risk extrapolation (rex).
\newblock In {\em ICML}, 2021.

\bibitem{lampert2009learning}
Christoph~H Lampert, Hannes Nickisch, and Stefan Harmeling.
\newblock Learning to detect unseen object classes by between-class attribute
  transfer.
\newblock In {\em CVPR}, pages 951--958. IEEE, 2009.

\bibitem{li2017deeper}
Da Li, Yongxin Yang, Yi-Zhe Song, and Timothy~M Hospedales.
\newblock Deeper, broader and artier domain generalization.
\newblock In {\em ICCV}, pages 5542--5550, 2017.

\bibitem{li2021metasaug}
Shuang Li, Kaixiong Gong, Chi~Harold Liu, Yulin Wang, Feng Qiao, and Xinjing
  Cheng.
\newblock Metasaug: Meta semantic augmentation for long-tailed visual
  recognition.
\newblock In {\em CVPR}, pages 5212--5221, 2021.

\bibitem{li2020symmetry}
Yong-Lu Li, Yue Xu, Xiaohan Mao, and Cewu Lu.
\newblock Symmetry and group in attribute-object compositions.
\newblock In {\em CVPR}, pages 11316--11325, 2020.

\bibitem{liu2018learning}
Yanbin Liu, Juho Lee, Minseop Park, Saehoon Kim, Eunho Yang, Sung~Ju Hwang, and
  Yi Yang.
\newblock Learning to propagate labels: Transductive propagation network for
  few-shot learning.
\newblock 2018.

\bibitem{liu2019openlongtailrecognition}
Ziwei Liu, Zhongqi Miao, Xiaohang Zhan, Jiayun Wang, Boqing Gong, and Stella~X.
  Yu.
\newblock Large-scale long-tailed recognition in an open world.
\newblock In {\em CVPR}, 2019.

\bibitem{mancini2021open}
Massimiliano Mancini, Muhammad~Ferjad Naeem, Yongqin Xian, and Zeynep Akata.
\newblock Open world compositional zero-shot learning.
\newblock In {\em CVPR}, pages 5222--5230, 2021.

\bibitem{mondal2021mini}
Arnab~Kumar Mondal, Vineet Jain, and Kaleem Siddiqi.
\newblock Mini-batch graphs for robust image classification.
\newblock {\em arXiv preprint arXiv:2105.03237}, 2021.

\bibitem{naeem2021learning}
MF Naeem, Y Xian, F Tombari, and Zeynep Akata.
\newblock Learning graph embeddings for compositional zero-shot learning.
\newblock In {\em CVPR}. IEEE, 2021.

\bibitem{narayan2020latent}
Sanath Narayan, Akshita Gupta, Fahad~Shahbaz Khan, Cees~GM Snoek, and Ling
  Shao.
\newblock Latent embedding feedback and discriminative features for zero-shot
  classification.
\newblock In {\em ECCV}, pages 479--495. Springer, 2020.

\bibitem{oord2018representation}
Aaron van~den Oord, Yazhe Li, and Oriol Vinyals.
\newblock Representation learning with contrastive predictive coding.
\newblock In {\em NeurIPS}, 2019.

\bibitem{palatucci2009zero}
Mark~M Palatucci, Dean~A Pomerleau, Geoffrey~E Hinton, and Tom Mitchell.
\newblock Zero-shot learning with semantic output codes.
\newblock 2009.

\bibitem{peng2019domain}
Xingchao Peng, Zijun Huang, Ximeng Sun, and Kate Saenko.
\newblock Domain agnostic learning with disentangled representations.
\newblock In {\em ICML}, pages 5102--5112. PMLR, 2019.

\bibitem{visda2017}
Xingchao Peng, Ben Usman, Neela Kaushik, Judy Hoffman, Dequan Wang, and Kate
  Saenko.
\newblock Visda: The visual domain adaptation challenge, 2017.

\bibitem{piratla2020efficient}
Vihari Piratla, Praneeth Netrapalli, and Sunita Sarawagi.
\newblock Efficient domain generalization via common-specific low-rank
  decomposition.
\newblock In {\em ICML}, pages 7728--7738. PMLR, 2020.

\bibitem{purushwalkam2019task}
Senthil Purushwalkam, Maximilian Nickel, Abhinav Gupta, and Marc'Aurelio
  Ranzato.
\newblock Task-driven modular networks for zero-shot compositional learning.
\newblock In {\em ICCV}, pages 3593--3602, 2019.

\bibitem{ren2020balanced}
Jiawei Ren, Cunjun Yu, Shunan Sheng, Xiao Ma, Haiyu Zhao, Shuai Yi, and
  Hongsheng Li.
\newblock Balanced meta-softmax for long-tailed visual recognition.
\newblock In {\em NeurIPS}, 2020.

\bibitem{selvaraju2017grad}
Ramprasaath~R Selvaraju, Michael Cogswell, Abhishek Das, Ramakrishna Vedantam,
  Devi Parikh, and Dhruv Batra.
\newblock Grad-cam: Visual explanations from deep networks via gradient-based
  localization.
\newblock In {\em ICCV}, pages 618--626, 2017.

\bibitem{shankar2018generalizing}
Shiv Shankar, Vihari Piratla, Soumen Chakrabarti, Siddhartha Chaudhuri, Preethi
  Jyothi, and Sunita Sarawagi.
\newblock Generalizing across domains via cross-gradient training.
\newblock In {\em ICLR}, 2018.

\bibitem{shen2020invertible}
Yuming Shen, Jie Qin, Lei Huang, Li Liu, Fan Zhu, and Ling Shao.
\newblock Invertible zero-shot recognition flows.
\newblock In {\em ECCV}, pages 614--631. Springer, 2020.

\bibitem{deep_coral}
Baochen Sun and Kate Saenko.
\newblock Deep coral: Correlation alignment for deep domain adaptation.
\newblock In {\em ECCV}, 2016.

\bibitem{tan2020equalization}
Jingru Tan, Changbao Wang, Buyu Li, Quanquan Li, Wanli Ouyang, Changqing Yin,
  and Junjie Yan.
\newblock Equalization loss for long-tailed object recognition.
\newblock In {\em CVPR}, pages 11662--11671, 2020.

\bibitem{tokmakov2019learning}
Pavel Tokmakov, Yu-Xiong Wang, and Martial Hebert.
\newblock Learning compositional representations for few-shot recognition.
\newblock In {\em CVPR}, pages 6372--6381, 2019.

\bibitem{touvron21adeit}
Hugo Touvron, Matthieu Cord, Matthijs Douze, Francisco Massa, Alexandre
  Sablayrolles, and Herve Jegou.
\newblock Training data-efficient image transformers \& distillation through
  attention.
\newblock In {\em ICML}, pages 10347--10357, 2021.

\bibitem{van2018inaturalist}
Grant Van~Horn, Oisin Mac~Aodha, Yang Song, Yin Cui, Chen Sun, Alex Shepard,
  Hartwig Adam, Pietro Perona, and Serge Belongie.
\newblock The inaturalist species classification and detection dataset.
\newblock In {\em CVPR}, pages 8769--8778, 2018.

\bibitem{vaswani2017attention}
Ashish Vaswani, Noam Shazeer, Niki Parmar, Jakob Uszkoreit, Llion Jones,
  Aidan~N Gomez, {\L}ukasz Kaiser, and Illia Polosukhin.
\newblock Attention is all you need.
\newblock In {\em NeurIPS}, pages 5998--6008, 2017.

\bibitem{venkateswara2017deep}
Hemanth Venkateswara, Jose Eusebio, Shayok Chakraborty, and Sethuraman
  Panchanathan.
\newblock Deep hashing network for unsupervised domain adaptation.
\newblock In {\em CVPR}, pages 5018--5027, 2017.

\bibitem{wah2011caltech}
Catherine Wah, Steve Branson, Peter Welinder, Pietro Perona, and Serge
  Belongie.
\newblock The caltech-ucsd birds-200-2011 dataset.
\newblock 2011.

\bibitem{wang2021generalizing}
Jindong Wang, Cuiling Lan, Chang Liu, Yidong Ouyang, Wenjun Zeng, and Tao Qin.
\newblock Generalizing to unseen domains: A survey on domain generalization.
\newblock {\em arXiv preprint arXiv:2103.03097}, 2021.

\bibitem{wang2021rsg}
Jianfeng Wang, Thomas Lukasiewicz, Xiaolin Hu, Jianfei Cai, and Zhenghua Xu.
\newblock Rsg: A simple but effective module for learning imbalanced datasets.
\newblock In {\em CVPR}, pages 3784--3793, 2021.

\bibitem{wang2021contrastive}
Peng Wang, Kai Han, Xiu-Shen Wei, Lei Zhang, and Lei Wang.
\newblock Contrastive learning based hybrid networks for long-tailed image
  classification.
\newblock In {\em CVPR}, pages 943--952, 2021.

\bibitem{wang2019survey}
Wei Wang, Vincent~W Zheng, Han Yu, and Chunyan Miao.
\newblock A survey of zero-shot learning: Settings, methods, and applications.
\newblock {\em ACM Transactions on Intelligent Systems and Technology (TIST)},
  10(2):1--37, 2019.

\bibitem{wang2020long}
Xudong Wang, Long Lian, Zhongqi Miao, Ziwei Liu, and Stella~X Yu.
\newblock Long-tailed recognition by routing diverse distribution-aware
  experts.
\newblock In {\em ICLR}, 2021.

\bibitem{wang2017learning}
Yu-Xiong Wang, Deva Ramanan, and Martial Hebert.
\newblock Learning to model the tail.
\newblock In {\em NIPS}, pages 7032--7042, 2017.

\bibitem{rw2019timm}
Ross Wightman.
\newblock Pytorch image models.
\newblock \url{https://github.com/rwightman/pytorch-image-models}, 2019.

\bibitem{wu2019detectron2}
Yuxin Wu, Alexander Kirillov, Francisco Massa, Wan-Yen Lo, and Ross Girshick.
\newblock Detectron2.
\newblock \url{https://github.com/facebookresearch/detectron2}, 2019.

\bibitem{xian2018zero}
Yongqin Xian, Christoph~H Lampert, Bernt Schiele, and Zeynep Akata.
\newblock Zero-shot learning—a comprehensive evaluation of the good, the bad
  and the ugly.
\newblock {\em PAMI}, 41(9):2251--2265, 2018.

\bibitem{xiang2020learning}
Liuyu Xiang, Guiguang Ding, and Jungong Han.
\newblock Learning from multiple experts: Self-paced knowledge distillation for
  long-tailed classification.
\newblock In {\em ECCV}, pages 247--263. Springer, 2020.

\bibitem{yu2014fine}
Aron Yu and Kristen Grauman.
\newblock Fine-grained visual comparisons with local learning.
\newblock In {\em CVPR}, pages 192--199, 2014.

\bibitem{zhang2018mixup}
Hongyi Zhang, Moustapha Cisse, Yann~N Dauphin, and David Lopez-Paz.
\newblock mixup: Beyond empirical risk minimization.
\newblock {\em ICLR}, 2018.

\bibitem{MDD}
Yuchen Zhang, Tianle Liu, Mingsheng Long, and Michael Jordan.
\newblock Bridging theory and algorithm for domain adaptation.
\newblock In {\em ICML}, 2019.

\bibitem{zhao2021learning}
Yuyang Zhao, Zhun Zhong, Fengxiang Yang, Zhiming Luo, Yaojin Lin, Shaozi Li,
  and Nicu Sebe.
\newblock Learning to generalize unseen domains via memory-based multi-source
  meta-learning for person re-identification.
\newblock In {\em CVPR}, pages 6277--6286, 2021.

\bibitem{zhong2021improving}
Zhisheng Zhong, Jiequan Cui, Shu Liu, and Jiaya Jia.
\newblock Improving calibration for long-tailed recognition.
\newblock In {\em CVPR}, pages 16489--16498, 2021.

\bibitem{zhou2020bbn}
Boyan Zhou, Quan Cui, Xiu-Shen Wei, and Zhao-Min Chen.
\newblock Bbn: Bilateral-branch network with cumulative learning for
  long-tailed visual recognition.
\newblock In {\em CVPR}, pages 9719--9728, 2020.

\bibitem{zhou2017places}
Bolei Zhou, Agata Lapedriza, Aditya Khosla, Aude Oliva, and Antonio Torralba.
\newblock Places: A 10 million image database for scene recognition.
\newblock {\em PAMI}, 40(6):1452--1464, 2017.

\bibitem{zhou2021domain}
Kaiyang Zhou, Ziwei Liu, Yu Qiao, Tao Xiang, and Chen~Change Loy.
\newblock Domain generalization: A survey.
\newblock {\em arXiv preprint arXiv:2103.02503}, 2021.

\bibitem{zhou2020learning}
Kaiyang Zhou, Yongxin Yang, Timothy Hospedales, and Tao Xiang.
\newblock Learning to generate novel domains for domain generalization.
\newblock In {\em ECCV}, pages 561--578. Springer, 2020.

\bibitem{zhou2021mixstyle}
Kaiyang Zhou, Yongxin Yang, Yu Qiao, and Tao Xiang.
\newblock Domain generalization with mixstyle.
\newblock {\em ICLR}, 2021.

\bibitem{zhu2020inflated}
Linchao Zhu and Yi Yang.
\newblock Inflated episodic memory with region self-attention for long-tailed
  visual recognition.
\newblock In {\em CVPR}, pages 4344--4353, 2020.

\bibitem{zhu2018generative}
Yizhe Zhu, Mohamed Elhoseiny, Bingchen Liu, Xi Peng, and Ahmed Elgammal.
\newblock A generative adversarial approach for zero-shot learning from noisy
  texts.
\newblock In {\em CVPR}, pages 1004--1013, 2018.

\end{thebibliography}
}

\appendix
\section{Additional Experimental Details}

In all our experiments, if not otherwise stated, we use one layer Transformer Encoder. Meanwhile, The proposed BatchFormer is inserted after the global average pooling layer in ResNet.

{\bf Long-Tailed Recognition}. For Balanced Softmax~\cite{ren2020balanced}, we directly insert BatchFormer before the classifier, and train the network with 128 batch size on 1 V100 GPU for 90 epochs. The initial learning rate for Balanced Softmax~\cite{ren2020balanced} is 0.05 (it is linearly decreased according to the batch size) and we set the learning rate of the weights of BatchFormer module to 0.005 to avoid overfitting. For RIDE~\cite{wang2020long}, we train the network with a batch size of 400 on 4 V100 GPUs for 100 epochs with an initial learning rare of 0.1 on ImageNet-LT and 0.2 on iNaturalist 2018 respectively. For a fair comparison, we verify BatchFormer based on 3 experts RIDE. Specifically, a shared BatchFormer is incorporated in all expert branches, \ie the module of BatchFormer is shared among 3 expert heads. Experimentally, we find it is not necessary to utilized shared classifier for RIDE. We thus remove shared classifier in RIDE. We think this might be because the multiple experts internally maintain the features between before and after BatchFormer. For Paco~\cite{cui2021parametric}, we use default hyper-parameters of Paco on ImageNet-LT with four V100 GPUs, on CIFAR-100-LT with a single V100 GPU, and on Places-LT with a single V100 GPU. We simply insert BatchFormer in the query encoder, momentum encoder and prediction head, respectively. Meanwhile, the proposed BatchFormer is shared among those places. See more details in the provided code for MoCo. We find it requires 9 days to train Paco on iNaturalist18~\cite{van2018inaturalist} with 4 V100 GPUs. Thus, we directly evaluate BatchFormer on RIDE which only requires 32 hours.

{\bf Compositional Zero-Shot Learning}. We would like to clarify some details when reproducing the results of  ~\cite{naeem2021learning} on UT-Zap50K. Specifically, we can not fully reproduce the result of ~\cite{naeem2021learning} on UT-Zap50K due to the missing details. We also find the result with learnable feature extractor of~\cite{naeem2021learning} on UT-Zap50K is much better than the result with fixed feature extractor in Table 2 of~\cite{naeem2021learning}, which is largely different from the results on other datasets. Meanwhile, the same problem has been noticed by others on github issues of  ~\cite{naeem2021learning}, i.e., the result can not be unable to fully reproduced (\hyperlink{https://github.com/ExplainableML/czsl/issues/4}{https://github.com/ExplainableML/czsl/issues/4}). Therefore, all our experiments are based on the reproduced result for a fair comparison with~\cite{naeem2021learning}.

{\bf Domain Generalization.} We mainly evaluate BatchFormer on domain generalization by using the following two baseline methods: Transfer-Learning-Library~\cite{dalib} and SWAD~\cite{cha2021swad}. ~\cite{dalib} provides massive traditional and recent methods for domain generalization, and we thus simply apply BatchFormer to those method and evaluate the effectiveness of BatchFormer. Specifically, we use the default setting of ~\cite{dalib} for each method, and reproduce the result of each method with~\cite{dalib} for a fair comparison. All experiments are conducted three times and the results are averaged. For SWAD~\cite{cha2021swad}, we use the released code and default setting to reproduce the result of ResNet-18.For fair comparison, we also provide the result of ResNet-50.

{\bf Self-Supervised Learning.} BatchFormer is also pluggable to contrastive learning. Specifically, with BatchFormer, we train MoCo-v2~\cite{chen2020mocov2} and MoCo-v3~\cite{chen2021mocov3} for 200 epochs and 300 epochs respectively. We follow all default training settings when comparsing with MoCo-v2 and MoCo-v3. Since it is different from supervised learning, we provide the code based on MoCo-v3. As the released code of MoCo requires 16 GPUs with 32GB memory, we conduct all contrastive learning experiments using two cluster nodes with 8 NVIDIA A100 GPUs (40GB) for each node. We keep all other experimental details same as~\cite{chen2020mocov2} and~\cite{chen2021mocov3} for a fair comparison.

\begin{table}[tp]
\small
\caption{Illustration of BatchFormer on Generalized Zero-Shot Learning based on ~\cite{wang2021contrastive}. Unseen and Seen are the Top-1 accuracies tested on unseen classes and seen classes, respectively, in GZSL. }
\label{table:gzsl}
\centering

\begin{tabular}{@{}lccc@{}}
\hline
 \multirow{2}{*}{Method} &
\multicolumn{3}{c}{CUB~\cite{wah2011caltech}}\cr
 & Unseen & Seen & Harmonic mean \\
 \hline
 IZF~\cite{shen2020invertible} &   52.7  &  {\bf 68.0} &   59.4 \\
 TF-VAEGAN~\cite{narayan2020latent} & 52.8 &  64.7  & 58.1 \\
CE-GZSL~\cite{wang2021contrastive} & 63.9 & 66.8 & 65.3 \\
\hline
CE-GZSL*(reproduced) & 67.5 & 65.1 & 66.3\\
+ BatchFormer & {\bf 68.2} & 65.8 & {\bf 67.0}\\
\hline
\end{tabular}
\end{table}



\section{Additional Experiments}
To better demonstrate the effectiveness of the proposed BatchFormer, we provide additional experimental results on more tasks.

\subsection{Generalized Zero-Shot Learning}

We also evaluate BatchFormer on generalized zero-shot learning task. Specifically, we report the accuracy of ``seen",  ``unseen", and  the harmonic mean of them (unseen and seen). We perform experiments on one of the most popular datasets for generalized zero-shot learning, CUB~\cite{wah2011caltech}, which includes 11,788 images from 200 bird species. We build a baseline with the released code of ~\cite{wang2021contrastive} and achieve better results than ~\cite{wang2021contrastive}. As shown in Table~\ref{table:gzsl}, the proposed BatchFormer achieves a new state-of-the-art on Unseen and Harmonic mean.



\subsection{Self-Supervised Learning}

{\bf Object detection on VOC2007}. We also evaluate Object Detection of MoCo on VOC2007~\cite{everingham2010pascal} in Table~\ref{table:moco_det}. Similar as MoCo-v2~\cite{chen2020mocov2}, we use the pre-trained model to fine-tune Faster-RCNN on VOC2007 based on Detectron2~\cite{wu2019detectron2}. We find MoCo-v3 achieves worse result on VOC2007. However, BatchFormer consistently improves the object detection on VOC2007. Here, we train MoCo-v2 for 200 epochs, and MoCo-v3 for 100 epochs. Specifically, we think that the number of training epochs (only 100 epochs) of MoCo-v3 might limit the performance on VOC2007.

\begin{table}[tp]
\small
\caption{Illustration of BatchFormer for MoCo on VOC2007~\cite{everingham2010pascal}. Here, for a fair comparison, we use the the released code of MoCo-v2 and MoCo-v3 to run the experiments in the same setting, and obtain the baseline.}
\label{table:moco_det}
\centering
\begin{tabular}{@{}lccc@{}}
\hline
Methods & AP & AP50 & AP75 \\
\hline
MoCo-v2*~\cite{chen2020mocov2} & 56.4 & 82.1 & 63.1\\
+BatchFormer & {\bf 56.7} & {\bf 82.0} & {\bf 63.6}\\
\hline
MoCo-v3~\cite{chen2020mocov2} & 46.6 & 78.2 & 48.9\\
MoCo-v3~\cite{chen2020mocov2} & {\bf 48.0} & {\bf 78.8} & {\bf 51.1}\\
\hline
\end{tabular}
\end{table}

\subsection{Image Recognition}
Table~\ref{table:recognition} demonstrates BatchFormer for Image Classification. We find BatchFormer achieves comparable performance among ResNet50. This shows BatchFormer does not degrade the performance when the distribution of data is balanced.

\begin{table}[tp]
\small
\caption{Illustration of BatchFormer for Image Recognition.}
\label{table:recognition}
\centering
\begin{tabular}{@{}lccc@{}}
\hline
Methods & Epochs & Top-1 & Top-5 \\
\hline
ResNet50~\cite{rw2019timm} & 200 & 78.9 &\\
ResNet50 + BatchFormer & 200 & 78.9 & -\\
\hline
\end{tabular}
\end{table}

\begin{table}[tp]
\setlength\tabcolsep{2.5pt}
\small
\caption{Illustration of BatchFormer for Domain Generalization under different works on PACS~\cite{li2017deeper}. Here, the baseline is from ~\cite{dalib}. SWAD~\cite{cha2021swad} is reproduced based on the released code of ~\cite{cha2021swad}.}
\label{table:dg_pacs_app}
\centering
\begin{tabular}{@{}lccccc@{}}
\hline
Methods  & art\_paint & cartoon & sketches & photo & Avg.  \\
\hline
Baseline & 81.3$\pm$0.7 & 76.1$\pm$0.6 & 75.5$\pm$2.6 & {\bf 95.4} $\pm$0.2 & 82.0 \\
+BatchFormer & {\bf 82.4}$\pm$1.5  & {\bf 76.4}$\pm$1.2  &  {\bf 75.7}$\pm$1.0 & 95.1$\pm$0.4 & {\bf 82.4} \\
\hline
CORAL~\cite{deep_coral} & 79.2$\pm$1.7 &  {\bf 75.5} $\pm$1.1 & 71.4$\pm$3.1 & 94.7$\pm$0.3 &  80.2 \\
+BatchFormer & {\bf 80.6}$\pm$0.9 &  74.7$\pm$1.9 & {\bf 73.1}$\pm$0.3& {\bf 95.1}$\pm$0.3 & {\bf 80.9} \\

\hline
IRM~\cite{arjovsky2019invariant} & {\bf 81.0}$\pm$0.6 &  {\bf 71.4}$\pm$4.1 & 68.1$\pm$7.1 & 95.0$\pm$0.6 &     78.9 \\
+BatchFormer & 78.9$\pm$3.1 &  71.0$\pm$7.1 &   {\bf 71.5}$\pm$2.8 & {\bf 96.0}$\pm$0.3   & {\bf 79.4}\\
\hline
V-REx~\cite{VREx} & 80.8$\pm$1.8 & 75.3$\pm$1.4 & 73.3$\pm$0.9 & 95.9$\pm$0.0 & 81.3 \\
+ BatchFormer & {\bf 82.0}$\pm$0.3 & {\bf 76.3}$\pm$0.7 & {\bf 75.2}$\pm$1.7 & 95.3$\pm$0.1 & {\bf 82.2} \\
\hline
MixStyle~\cite{zhou2021mixstyle} & 81.7$\pm$0.1 &  {\bf 76.8}$\pm$0.0 & 80.8$\pm$0.0 &  93.1$\pm$0.0 & 83.1 \\
+BatchFormer & {\bf 84.8} $\pm$0.4 &  75.3$\pm$0.0 &  {\bf 81.1} $\pm$0.4 & {\bf 93.6}$\pm$0.0 &  {\bf 83.7}\\
\hline
SWAD*~\cite{cha2021swad} & 83.1$\pm$1.5 & 75.9$\pm$0.9 &  77.1$\pm$2.4 & 95.6$\pm$0.6 & 82.9\\
+BatchFormer& {\bf 84.3}$\pm$0.8 & {\bf 76.9} $\pm$1.2 & {\bf 78.2}$\pm$1.8 & 95.7$\pm$0.6 &  {\bf 83.9}\\
\hline\hline
ResNet50 \\
\hline\hline
V-REx~\cite{VREx}     &  83.8$\pm$4.8& {\bf 81.0}$\pm$0.0 & {\bf 97.7}$\pm$0.4 &   77.7$\pm$3.1   & 85.0 \\
+ BatchFormer     &  {\bf 87.3} $\pm$5.0  &  80.2$\pm$4.6 &  97.1$\pm$1.7 &   {\bf 77.9}$\pm$4.4   & {\bf 85.6} \\
\hline
IRM~\cite{arjovsky2019invariant} & 88.2$\pm$0.6  & 79.8$\pm$1.0 &97.6$\pm$0.5 & 77.6$\pm$0.7 & 85.8 \\
+ BatchFormer & {\bf 89.0}$\pm$0.98 & {\bf  80.1 }$\pm$1.0 &{\bf 98.0}$\pm$0.4 & {\bf 79.8}$\pm$0.4 & {\bf 86.8} \\
\hline
SWAD~\cite{cha2021swad} & 89.4$\pm$0.7 & 83.7$\pm$1.2 & {\bf 97.7}$\pm$0.6 & 82.5$\pm$0.8 & 88.1\\
+BatchFormer& {\bf 90.2}$\pm$0.5 & {\bf 84.0}$\pm$1.0 & 97.3$\pm$0.3 & {\bf 83.0}$\pm$0.6 & {\bf 88.6}\\
\hline
\end{tabular}
\end{table}

\begin{figure*}
    \centering
    \includegraphics[width=.88\textwidth]{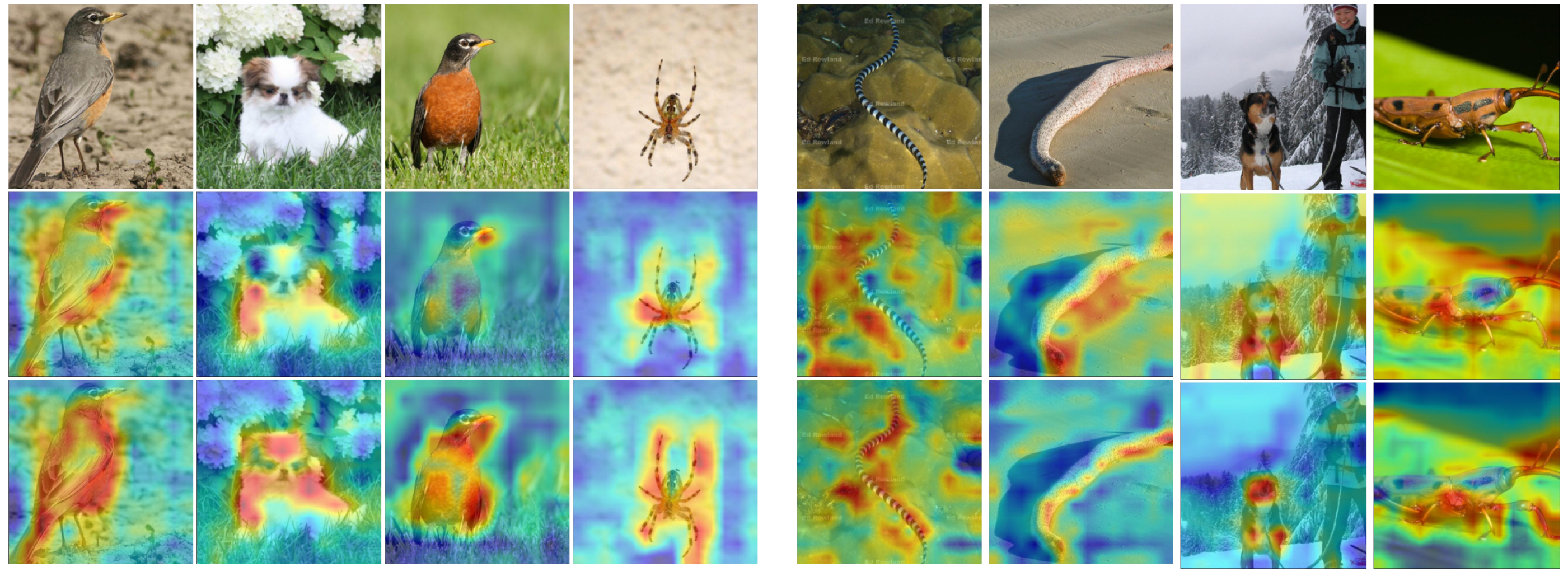}
    \caption{Grad-Cam demonstration of BatchFormer on low-shot test images based on~\cite{ren2020balanced}. The left images show BatchFormer enables the model pay attention on more details when the scene is simple, while the right images show BatchFormer facilitates the model ignore the spurious correlation in the image. This is clear version of Figure 5 in the paper.}
    \label{fig:cam_illu_app}
\end{figure*}

\subsection{Domain Generalization}

We provide more experimental results based on ~\cite{dalib} in Table~\ref{table:dg_pacs_app}. Experiments on OfficeHome, VLCS, TerraIncognita are provided in Table~\ref{table:dg_swad_officehome}, Table~\ref{table:dg_swad_vlcs} and Table~\ref{table:dg_swad_terra} respectively. The default backbone is ResNet-18.

\begin{table}[tp]
\small
\setlength\tabcolsep{2.5pt}
\caption{Illustration of BatchFormer for Domain Generalization based on recent work~\cite{cha2021swad} on OfficeHome }
\label{table:dg_swad_officehome}
\centering
\begin{tabular}{@{}lccccc@{}}
\hline
Methods  &  Art   & Clipart   & Product   & RealWorld & Avg.  \\
\hline
SWAD*~\cite{cha2021swad} & 54.5$\pm$0.8 & 49.4$\pm$0.1 & 70.9$\pm$0.1 & 72.7$\pm$0.2 & 62.1 \\
 + BatchFormer & {\bf 57.8}$\pm$0.1 & {\bf 51.0}$\pm$0.1 & {\bf 73.4}$\pm$0.2 & {\bf 75.1}$\pm$0.1 & {\bf 64.3}\\
\hline\hline
ResNet-50 \\
\hline\hline
IRM~\cite{arjovsky2019invariant} & 66.8$\pm$0.2  &  54.9$\pm$0.8  &  77.5$\pm$0.7 & 80.5$\pm$0.4  & 69.9 \\
+BatchFormer    & {\bf 67.7} $\pm$0.2  & {\bf 55.5}$\pm$0.8  & {\bf 78.4}$\pm$0.5 & {\bf 81.0}$\pm$0.3  & {\bf 70.6}\\

\hline
SWAD*~\cite{cha2021swad} & 65.9$\pm$0.8 & 58.0$\pm$0.1 & 78.5$\pm$0.5 & 80.2$\pm$0.7 & 70.6\\
 + BatchFormer & {\bf 66.7}$\pm$0.3 & 57.9$\pm$0.3 & {\bf 79.2}$\pm$0.4 & {\bf 80.6}$\pm$0.7 & {\bf 71.1}\\

\hline
\end{tabular}
\end{table}

\begin{table}[tp]
\small
\setlength\tabcolsep{2.5pt}
\caption{Illustration of BatchFormer for Domain Generalization based on recent work~\cite{cha2021swad} (ResNet-18) on VLCS }
\label{table:dg_swad_vlcs}
\centering
\begin{tabular}{@{}lccccc@{}}
\hline
Methods  & Caltech101 & LabelMe & SUN09 & SUN09 & Avg.  \\
\hline
SWAD*~\cite{cha2021swad} & 97.2$\pm$1.4 & 61.4$\pm$0.1 & 71.2$\pm$1.7 & 75.5$\pm$0.8 & 76.3  \\
 + BatchFormer & 97.2$\pm$0.8 & 61.3$\pm$1.1 & {\bf 71.7}$\pm$1.0 & {\bf 77.4}$\pm$0.4 & {\bf 76.9} \\

\hline
\end{tabular}
\end{table}

\begin{table}[tp]
\small
\setlength\tabcolsep{2.5pt}
\caption{Illustration of BatchFormer for Domain Generalization based on recent work~\cite{cha2021swad} (ResNet-18) on TerraIncognita }
\label{table:dg_swad_terra}
\centering
\begin{tabular}{@{}lccccc@{}}
\hline
Methods  &  Art   & Clipart   & Product   & RealWorld & Avg.  \\
\hline
SWAD*~\cite{cha2021swad} & 47.6$\pm$3.0 & 33.8$\pm$4.5 & 53.6$\pm$1.8 & 33.3$\pm$0.6 & 42.1 \\
 + BatchFormer & {\bf 49.8}$\pm$1.8 & {\bf 40.3}$\pm$2.0 & {\bf 55.2}$\pm$1.2 & {\bf 34.0}$\pm$1.1 & {\bf 44.8}\\

\hline
\end{tabular}
\end{table}

\subsection{Domain Adaption}

We also demonstrate BatchFormer on Domain Adaption on VisDA2017~\cite{visda2017}. Table~\ref{table:da_visda2017} shows BatchFormer effectively improves the corresponding baseline, \ie, MDD~\cite{MDD}.

\begin{table}[tp]
\small

\caption{Illustration of BatchFormer for Domain Adaption on VisDA2017~\cite{visda2017}. The backbone is ResNet-101. Experiments are based on ~\cite{dalib}. }
\label{table:da_visda2017}
\centering
\begin{tabular}{@{}lc@{}}
\hline
Methods  & Synthetic $->$ Real \\
\hline
MDD~\cite{MDD} & 76.8$\pm$1.5 \\
+BatchFormer & {\bf 77.8} $\pm$2.0 \\
\hline
\end{tabular}
\end{table}

\section{Ablation Studies}

\begin{table}[tp]
\small
\caption{Ablation Studies of different layers on ImageNet-LT based on Balanced Softmax~\cite{ren2020balanced}. The backbone is ResNet-10. }
\label{table:ab_modules_app}
\centering

\begin{tabular}{@{}lcccc@{}}
\hline
Method & All & Many & Medium & Few \\
\hline
BatchFormer (1 layers) & 43.2 & 52.8   &  40.4   &  25.6    \\
\hline

2  layers & 43.2  &52.8  &   40.3 &   26.0     \\
4 layers & 42.7& 52.1    & 40.1   & 25.4     \\
8 layers & 43.3 &53.3    & 40.2   & 26.1     \\
16 layers & 43.1 &  52.9  &   40.0 &   26.6       \\
\hline
\end{tabular}
\end{table}



\begin{table}[!bp]
\setlength\tabcolsep{2.pt}
\small
\caption{Illustration of BatchFormer without PaCo loss~\cite{cui2021parametric} on CIFAR-LT-100. }
\begin{center}
\label{table:cifar100_lt_ab_paco}
\centering

\begin{tabular}{@{}lcccc|cccc@{}}
\hline
 \multirow{2}{*}{Method} &
\multicolumn{4}{c}{100}&\multicolumn{4}{c}{200} \cr
 & All & Many & Med & Few  & All & Many & Med & Few \\
\hline
\hline
Baseline [52] & 52.0  & 68.1 & 53.2 &  31.6 &  47.31 & 67.8& 52.6 & 27.3 \\
+ BatchFormer & {\bf 52.6} & {\bf 68.7} & 53.2 & {\bf 33.1} & {\bf 48.13} & {\bf 68.9} & {\bf 53.1} & {\bf 28.2} \\
\hline
\end{tabular}
\end{center}
\end{table}

{\bf Number of Layers.} We use one layer BatchFormer in our experiment. Table~\ref{table:ab_modules_app} demonstrates with more layers of BatchFormer, we do not observe larger improvement. We leave how to leverage more layers of BatchFormer to improve the performance in future work.

{\bf BatchFormer without PaCo loss~\cite{cui2021parametric}.} We notice BatchFormer mainly improves PaCo on Many category on CIFAR-100-LT (imbalance ratio 100). We thus conduct additional ablation study on BatchFormer for PaCo. Here, we remove the PaCo loss with Balanced loss~\cite{cui2021parametric} to build the baseline. we observe consistent results in the Table~\ref{table:cifar100_lt_ab_paco}. Noticeably, the baseline without PaCo loss is even better than the one in the main paper.

\section{Visualized Comparison}
\begin{figure*}
    \centering
    \includegraphics[width=.99\textwidth]{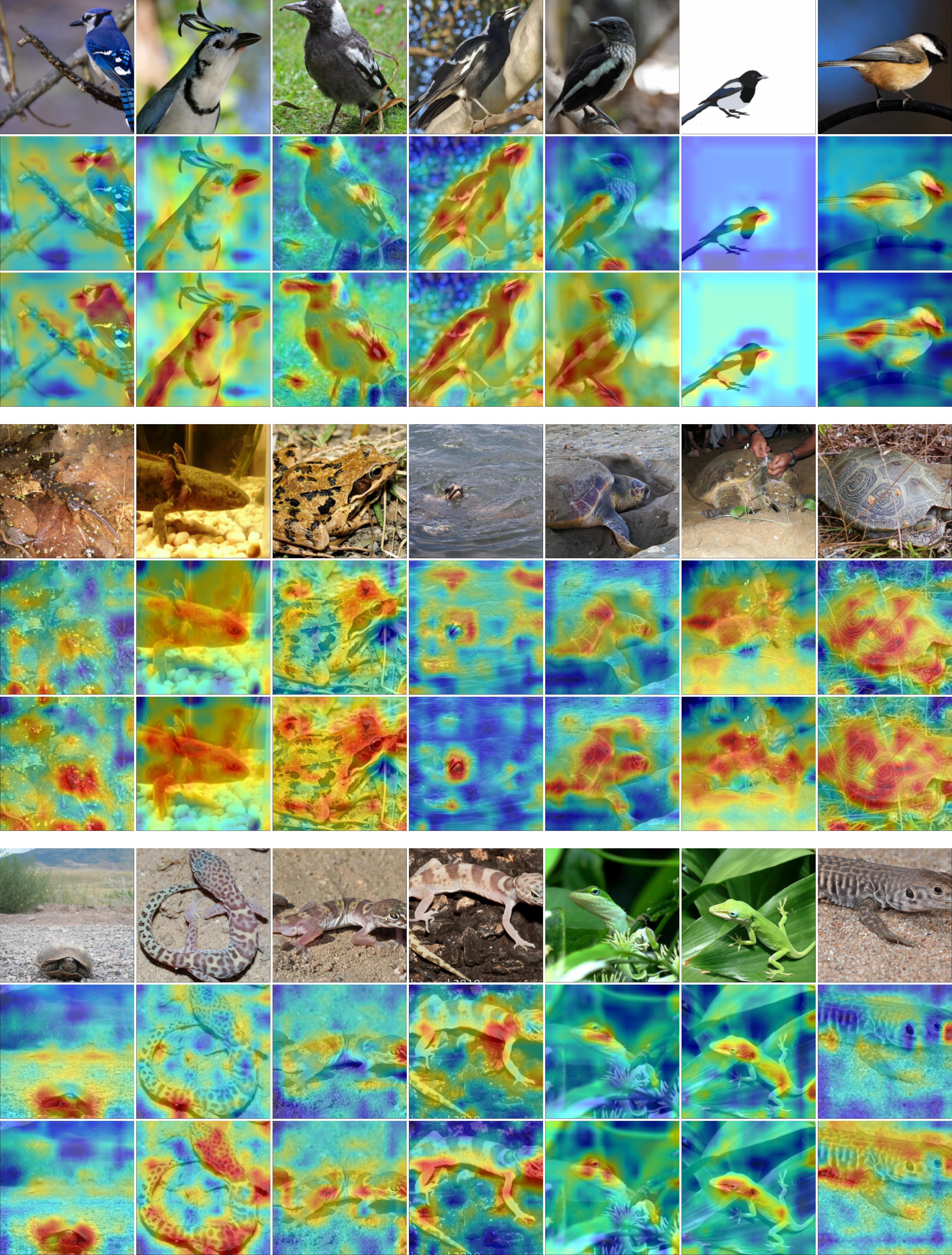}
    \caption{More Grad-Cam illustration of BatchFormer on low-shot test images based on~\cite{ren2020balanced}. The figures are also provided in the directory. }
    \label{fig:cam_illu2}
\end{figure*}

\begin{figure*}
    \centering
    \includegraphics[width=.99\textwidth]{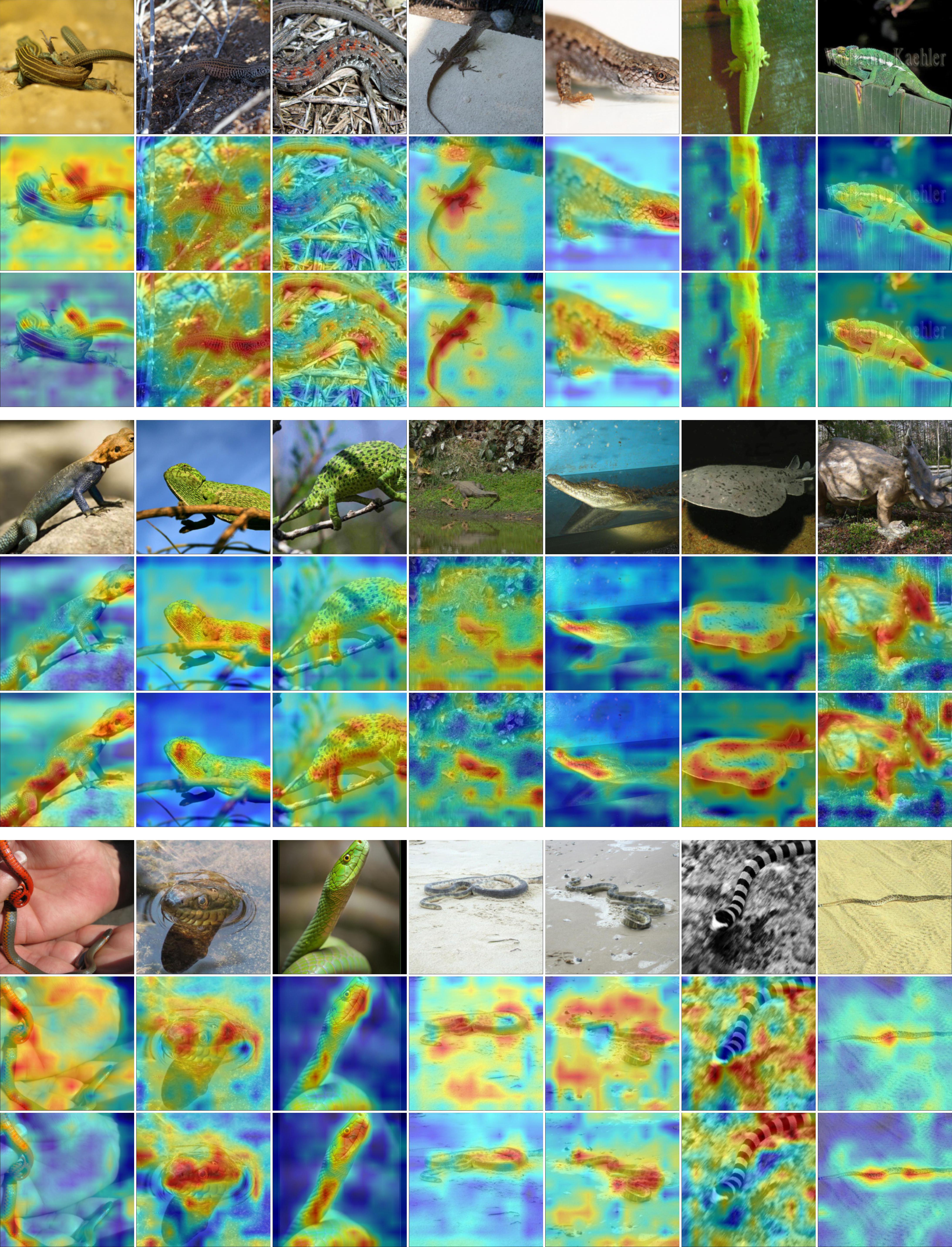}
    \caption{More Grad-Cam illustration of BatchFormer on low-shot test images based on~\cite{ren2020balanced}. The figures are also provided in the directory. }
    \label{fig:cam_illu3}
\end{figure*}

\begin{figure*}
    \centering
    \includegraphics[width=.99\textwidth]{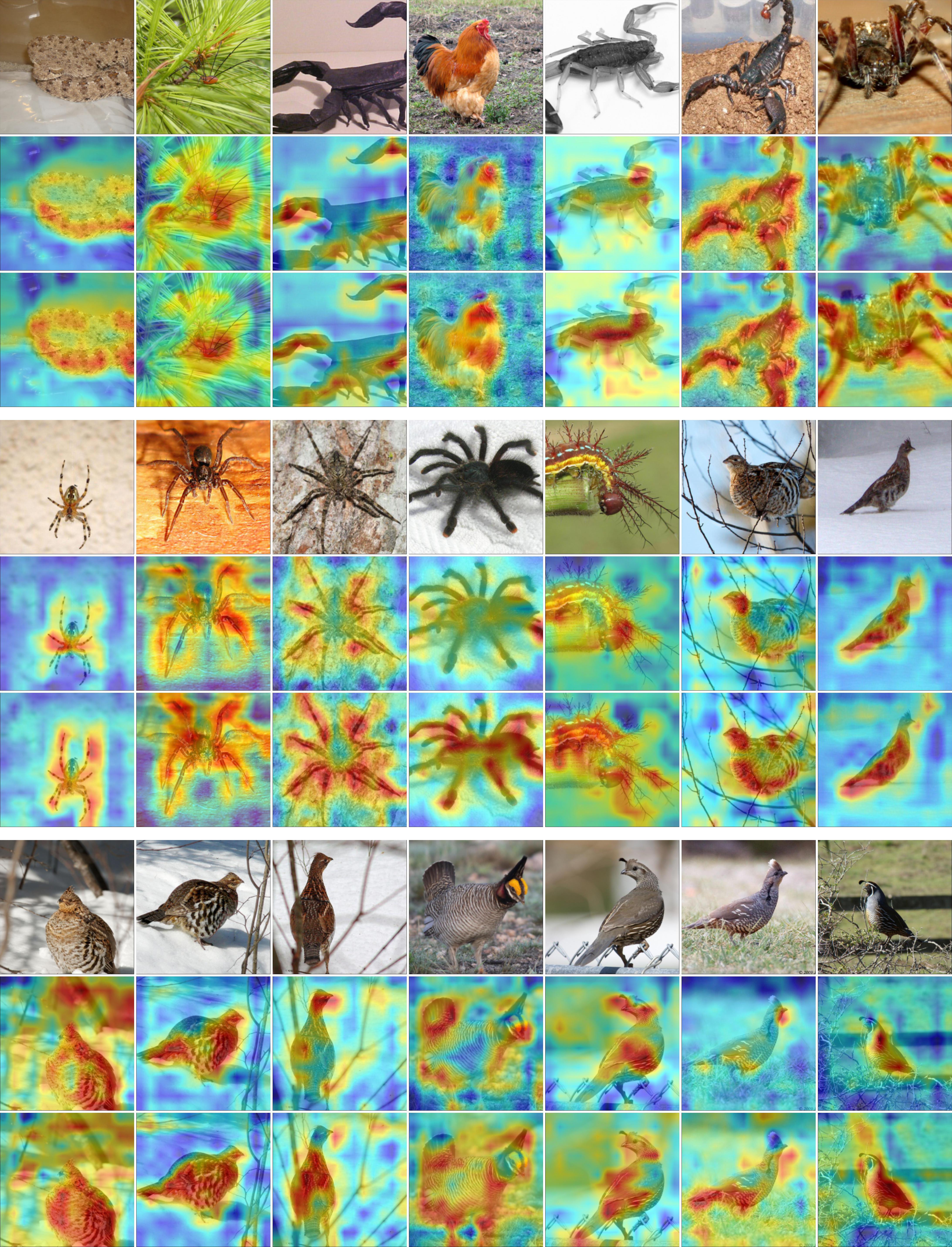}
    \caption{More Grad-Cam illustration of BatchFormer on low-shot test images based on~\cite{ren2020balanced}. The figures are also provided in the directory. }
    \label{fig:cam_illu4}
\end{figure*}

Figure~\ref{fig:cam_illu_app} provides clear figure of the Grad-Cam Figure in the main paper. More comparisons are included in Figure~\ref{fig:cam_illu2}, Figure~\ref{fig:cam_illu3}, Figure~\ref{fig:cam_illu4}, where we choose the top 100 classes on ImageNet for demonstration.

\clearpage


\end{document}